\documentclass[fleqn,10pt]{wlscirep}
\usepackage[utf8]{inputenc}
\usepackage[T1]{fontenc}
\usepackage{bm}
\usepackage{jabbrv}

\usepackage{graphicx}%
\usepackage{tabularx}
\usepackage{multirow}%
\usepackage{amsmath,amssymb,amsfonts}%
\usepackage{amsthm}%
\usepackage{mathrsfs}%
\usepackage[title]{appendix}%
\usepackage{xcolor}%
\usepackage{textcomp}%
\usepackage{manyfoot}%
\usepackage{booktabs}%
\usepackage{algorithm}%
\usepackage{algorithmicx}%
\usepackage{algpseudocode}%
\usepackage{listings}%
\usepackage{bbding}

\usepackage{adjustbox}
\usepackage{url}
\usepackage[left]{lineno}

\usepackage{pifont}
\usepackage{makecell}
\usepackage{geometry}
\usepackage{subcaption}
\usepackage{rotating}
\usepackage{xurl}
\usepackage{hyperref}
\usepackage{array}
\usepackage{xspace}
\usepackage{colortbl}
\usepackage{tcolorbox}
\definecolor{dino}{RGB}{249,231,227}
\definecolor{part}{RGB}{173,216,230}
\definecolor{all}{RGB}{135,206,235}

\title{Large-scale Self-supervised Video Foundation Model for Intelligent Surgery}

\author[1]{Shu Yang}

\author[1]{Fengtao Zhou}

\author[2,3]{Leon Mayer}

\author[1]{Fuxiang Huang}

\author[4]{Yiliang Chen}

\author[1]{Yihui Wang}

\author[1]{Sunan He}

\author[1]{Yuxiang Nie}

\author[1]{Xi Wang}

\author[2]{Ömer Sümer}

\author[5,6]{Yueming Jin}

\author[7]{Huihui Sun}

\author[7]{Shuchang Xu}

\author[8]{Alex Qinyang Liu}

\author[8]{Zheng Li}

\author[4]{Jing Qin}

\author[8]{Jeremy YuenChun Teoh}

\author[2,3,9,10,11, *]{Lena Maier-Hein}

\author[1,12,13,14,15, *]{Hao Chen}

\affil[1]{Department of Computer Science and Engineering, The Hong Kong University of Science and Technology, Hong Kong SAR, China}

\affil[2]{Division of Intelligent Medical Systems, German Cancer Research Center (DKFZ) Heidelberg, Heidelberg, Germany}

\affil[3]{Faculty of Medicine, Heidelberg University Hospital, Heidelberg, Germany}

\affil[4]{School of Nursing, The Hong Kong Polytechnic University, Hong Kong SAR, China}

\affil[5]{Department of Biomedical Engineering, National University of Singapore, Singapore}

\affil[6]{Department of Electrical and Computer Engineering, National University of Singapore, Singapore}

\affil[7]{Department of Gastroenterology of Tongji Hospital, School of Medicine, Tongji University, Shanghai, China}

\affil[8]{Department of Surgery, The Chinese University of Hong Kong, Hong Kong SAR, China}

\affil[9]{HI Helmholtz Imaging, German Cancer Research Center (DKFZ) Heidelberg, Heidelberg, Germany}

\affil[10]{Faculty of Mathematics and Computer Science, Heidelberg University, Heidelberg, Germany}

\affil[11]{National Center for Tumor Diseases (NCT), NCT Heidelberg, Heidelberg, Germany}

\affil[12]{Department of Chemical and Biological Engineering, The Hong Kong University of Science and Technology, Hong Kong SAR, China}

\affil[13]{Division of Life Science, The Hong Kong University of Science and Technology, Hong Kong SAR, China}

\affil[14]{State Key Laboratory of Molecular Neuroscience, The Hong Kong University of Science and Technology, Hong Kong SAR, China}

\affil[15]{Shenzhen-Hong Kong Collaborative Innovation Research Institute, The Hong Kong University of Science and Technology, Shenzhen, China}

\affil[*]{e-mail: jhc@cse.ust.hk}

\begin{abstract}
    
Computer-Assisted Intervention (CAI) has the potential to revolutionize modern surgery, with surgical scene understanding serving as a critical component in supporting decision-making, improving procedural efficacy, and ensuring intraoperative safety. While existing AI-driven approaches alleviate annotation burdens via self-supervised spatial representation learning, their lack of explicit temporal modeling during pre-training fundamentally restricts the capture of dynamic surgical contexts, resulting in incomplete spatiotemporal understanding. In this work, we introduce the first video-level surgical pre-training framework that enables joint spatiotemporal representation learning from large-scale surgical video data. To achieve this, we constructed a large-scale surgical video dataset comprising 3,650 videos and approximately 3.55 million frames, spanning more than 20 surgical procedures and over 10 anatomical structures. Building upon this dataset, we propose \textbf{SurgVISTA} (\textbf{Surg}ical \textbf{Vi}deo-level \textbf{S}patial-\textbf{T}emporal \textbf{A}rchitecture), a reconstruction-based pre-training method that captures intricate spatial structures and temporal dynamics through joint spatiotemporal modeling. Additionally, SurgVISTA incorporates image-level knowledge distillation guided by a surgery-specific expert to enhance the learning of fine-grained anatomical and semantic features. To validate its effectiveness, we established a comprehensive benchmark comprising 13 video-level datasets spanning six surgical procedures across four tasks. Extensive experiments demonstrate that SurgVISTA consistently outperforms both natural- and surgical-domain pre-trained models, demonstrating strong potential to advance intelligent surgical systems in clinically meaningful scenarios.

\end{abstract}

\begin{document}

\flushbottom
\maketitle

\section{Introduction}\label{sec1}

Computer-Assisted Intervention (CAI) has recently made significant strides in integrating sophisticated artificial intelligence (AI) technologies into clinical workflows to optimize preoperative planning, intraoperative execution, and postoperative assessment. A fundamental component of CAI is surgical scene understanding, which involves the comprehensive analysis and interpretation of intricate surgical activities and tissue interactions to support informed decision-making, improve procedural effectiveness, and ensure operational safety~\cite{kiyasseh2023vision,ma2022surgical,kiyasseh2023human}. This interdisciplinary endeavor aims to provide critical insights through continuous monitoring of intraoperative workflows and procedural execution, thereby facilitating surgical process optimization and promoting improved patient outcomes. However, developing AI-driven approaches for comprehensive surgical scene understanding encounters a significant challenge: the heavy reliance on expert-annotated surgical datasets. The SNOMED-CT International Edition (April 2025) catalogues thousands of procedures that differ markedly in prevalence and technical complexity.  Collecting adequately annotated video data for this long‑tail distribution is prohibitively expensive and labour‑intensive, resulting in surgical datasets that cover only a limited and incomplete spectrum of procedures. Consequently, the restricted procedural scope of existing surgical datasets hampers model generalizability, highlighting the inadequacy of task-specific and narrow AI in addressing the heterogeneous demands of surgical practice~\cite{workflowsurvey}, and emphasizing the need for procedure-agnostic methodologies capable of generalizing across diverse clinical contexts. In fact, the international Surgical Data Science Initiative~\cite{maier2017surgical,sds_website} has recently identified the methodological addressing of surgical data sparsity as a key next step in the field~\cite{maier2022surgical}. 

\begin{figure}[!ht]
\centering
\includegraphics[width=\linewidth]{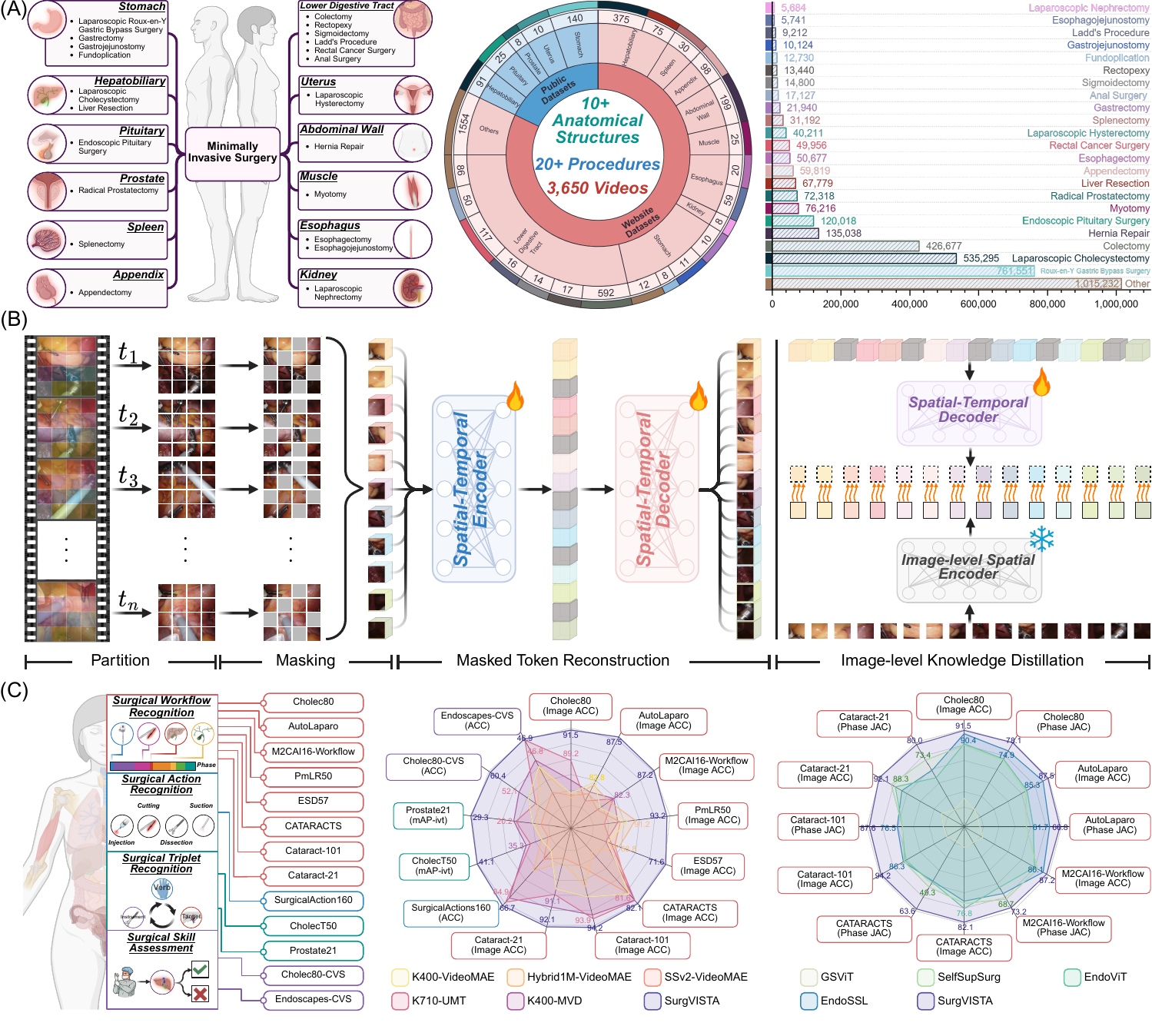}
\caption{\textbf{Overview of the study}. (A) \textbf{Pre-training Dataset}: Illustration of the primary anatomical structures and their associated surgical procedures, complemented by video-level statistics and detailed frame-level distribution analysis. (B) \textbf{SurgVISTA Framework}: An asymmetric encoder-decoder architecture featuring a unified encoder for comprehensive spatiotemporal modeling, and two decoders: one dedicated to video-level reconstruction for capturing temporal dynamics and structural consistency, and another for image-level knowledge distillation guided by an expert model to enhance spatial representations. (C) \textbf{Downstream Evaluation}: 
Comprehensive visualization of the four downstream tasks along with 13 evaluated datasets, accompanied by comparative performance analyses of models
pre-trained on natural-domain and surgical-domain data.}
\label{overview}
\end{figure}

Inspired by the rapid advancements in self-supervised learning (SSL) paradigm within the natural domain~\cite{DINO,MAE,dinov2} and its demonstrated effectiveness across various medical disciplines~\cite{azizi2023robust,ma2024towards,huang2023self,krishnan2022self,chen2024towards}, leveraging large-scale surgical data for SSL-based pre-training has emerged as a promising direction to extract robust and transferable representations explicitly tailored to surgical contexts. Despite significant achievements of SSL in various medical fields such as radiology~\cite{azizi2023robust,huang2023self} and pathology~\cite{ma2024towards,chen2024towards}, its application within the surgical field remains nascent and underexplored. Recent works~\cite{selfsupsurg, endovit, endossl, GSViT} have begun to specifically explore the SSL paradigm in the surgical domain, demonstrating initial successes in developing surgical foundation models and substantially improving surgical feature representation learning for downstream tasks. However, these methods predominantly focus on image-level pre-training for effective spatial representation learning, while neglecting the learning of temporal dynamics essential for comprehensive spatiotemporal representations. Consequently, they follow a mainstream two-stage paradigm~\cite{TeCNO}, where temporal dependencies are modeled during fine-tuning using external temporal modules applied to static and pre-extracted spatial features, resulting in a decoupled modeling process that hinders joint spatiotemporal representation learning. In practice, this decoupling is suboptimal for surgical video analysis, where seamless integration of spatial and temporal information is critical to address inherent complexities such as visual ambiguities from occlusions, motion blur, and procedure-specific variability~\cite{workflowsurvey}. Furthermore, a significant discrepancy exists between the upstream pre-training and downstream fine-tuning, as these methods adopt image-level pre-training combined with a two-stage fine-tuning framework. This fragmented pipeline hinders the unified learning of spatial and temporal representations: while spatial features benefit from a well pre-trained backbone, temporal modeling starts from a randomly initialized module. As a result, image-level  pre-trained methods struggle to leverage the rich temporal dynamics embedded in large-scale surgical video data, instead depending on limited downstream annotations to learn temporal dependencies. This piecewise optimization potentially compromises the robustness of spatiotemporal modeling, hindering generalizability across diverse downstream applications and resulting in suboptimal performance. These limitations highlight the urgent need for new approaches that holistically integrate spatial and temporal learning during pre-training to enhance the comprehensive surgical scene understanding.

This study advocates for explicit and unified spatiotemporal representation learning during the pre-training phase to fully exploit the potential of large-scale surgical video data, thereby initiating a paradigm shift from static image-based modeling to dynamic video-level representation learning. Such integration effectively facilitates the extraction of robust, transferable, and discriminative spatiotemporal feature representations, thereby advancing comprehensive surgical scene understanding. To support this objective, we constructed a publicly available, large-scale surgical video dataset comprising two essential components: SurgPub and SurgWeb, which collectively span varied anatomical structures and surgical procedures, providing the variability and complexity necessary for robust spatiotemporal representation learning. In Fig.~\ref{overview} (A), we present an overview of the key anatomical structures and associated surgical procedures, accompanied by video-level statistical analyses and frame-level distribution analyses. SurgPub is constructed from eight publicly available surgical video datasets~\cite{endonet,m2cai16workflow,HeiChole,PitVis,PSI-AVA,AutoLaparo,MultiBypass140}, encompassing 274 videos with a total of 1.20 million frames. Complementing SurgPub, SurgWeb comprises 3,376 surgical recordings sourced from online platforms, totaling 2.35 million frames, spanning diverse surgical procedures and anatomical regions encountered in extensive clinical scenarios. The resulting dataset encompasses over 20 surgical procedures across more than 10 anatomical structures, thereby promoting the learning of robust and generalizable spatiotemporal representations. Furthermore, we envision this curated resource as a catalyst for advancing more robust and generalized surgery AI models, while empowering researchers to tackle complex challenges in surgical scene understanding.

To transcend static image-based pre-training paradigms and establish video-level representation learning as the foundation for surgical AI, we pioneer \textbf{SurgVISTA} (\textbf{Surg}ical \textbf{Vi}deo-level \textbf{S}patial-\textbf{T}emporal \textbf{A}rchitecture), the first surgical video foundation model explicitly designed to capture robust and discriminative spatiotemporal patterns across diverse surgical scenarios. Unlike conventional methods restricted to static frame-level representations, SurgVISTA integrates spatial and temporal modeling into a unified pre-training framework, simultaneously encoding fine-grained anatomical details and temporal dynamics. As depicted in Fig.~\ref{overview} (B), SurgVISTA employs an asymmetric encoder-decoder architecture, comprising a unified encoder for joint spatiotemporal dependencies modeling and two dedicated decoders. The reconstruction decoder learns to restore masked video regions, thereby compelling the model to extract spatial structures and temporal dynamics intrinsic to surgical activities. Concurrently, the refinement decoder leverages knowledge distillation from a powerful image-level expert model~\cite{endossl}, compensating for detail loss from temporal abstraction while simultaneously reinforcing the discriminative capacity of spatial features. Through joint spatiotemporal modeling on large-scale surgical video data, SurgVISTA acquires highly generalizable representations that effectively encapsulate the complexity and variability of surgical scenarios, enabling seamless adaptation to downstream tasks without additional temporal components, thus significantly bridging the gap between pre-training and fine-tuning phases. Overall, SurgVISTA establishes a scalable and pretraining-driven foundation for surgical scene understanding, significantly enhancing robust spatiotemporal modeling capabilities for complex intraoperative scenarios.

To verify the effectiveness of SurgVISTA, we conducted a comprehensive video-level evaluation across 13 datasets, covering six surgical procedures and four distinct downstream tasks. As illustrated in Fig.~\ref{overview} (C), we provide a visual abstraction of diverse downstream tasks alongside the evaluated datasets, accompanied by comparative performance analyses of models pre-trained on natural-domain and surgical-domain data. Specifically, we began by comparing image-level and video-level pre-trained models within the natural domain, highlighting the superior performance of SurgVISTA and underscoring the necessity of domain-aligned video-level pre-training. Furthermore, we benchmarked SurgVISTA against the existing surgical foundation models, with a particular focus on surgical workflow recognition, demonstrating the advantages of joint spatiotemporal modeling empowered by large-scale surgical video pre-training in enhancing downstream performance. Despite the absence of task-specific data during the pre-training phase, SurgVISTA demonstrates exceptional generalization capabilities on anatomically and procedurally divergent surgical datasets, reinforcing its robustness and adaptability to a wide spectrum of surgical scenarios.
\section{Results}\label{sec2}

\begin{figure}[!h]
\centering
\includegraphics[width=0.88\linewidth, page=1]{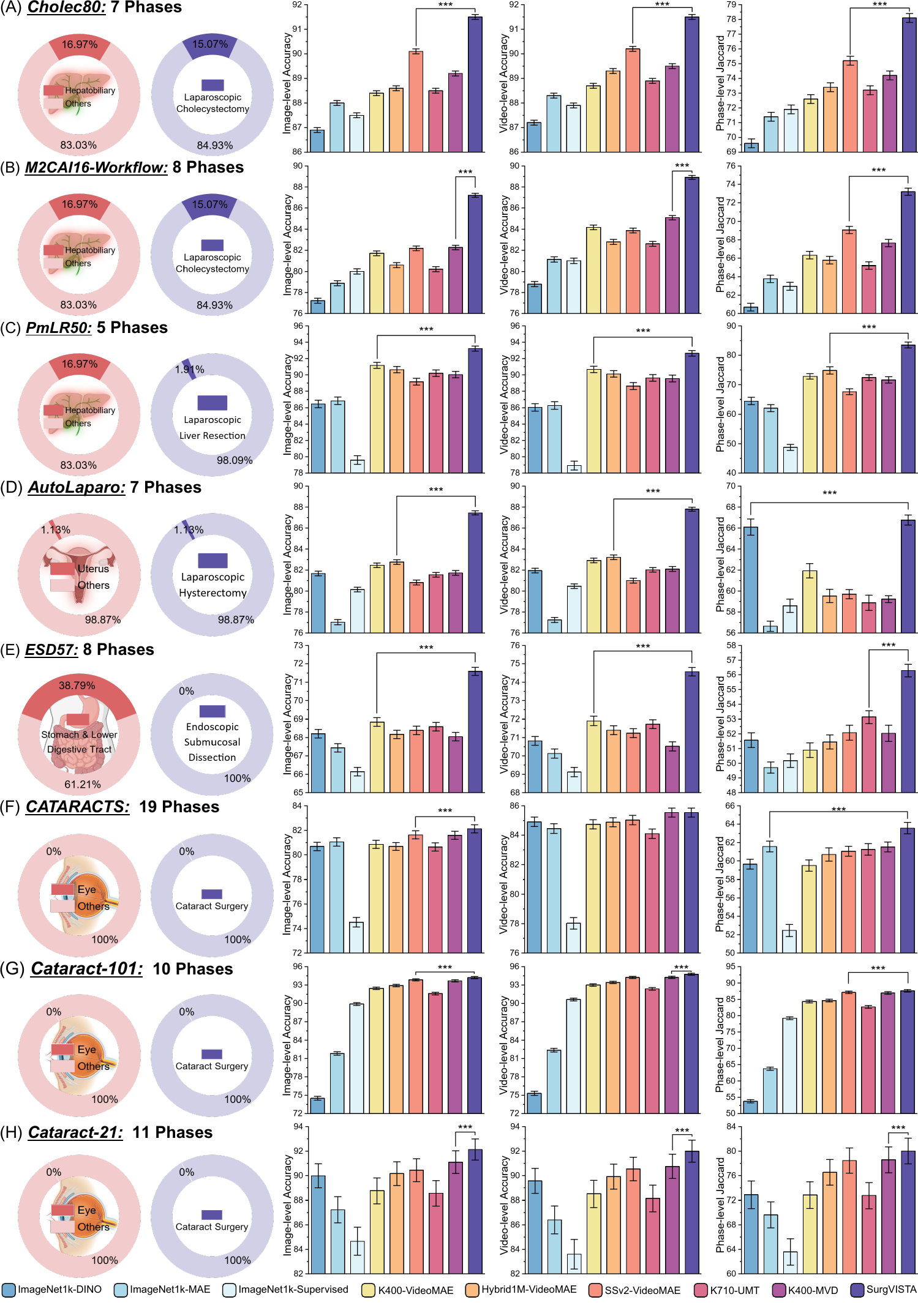}
\end{figure}

\begin{figure}[!h]
\caption{Experimental results of various natural-domain pre-trained methods and SurgVISTA on surgical workflow recognition datasets. Subfigures (A)-(D) correspond to in-domain datasets, while (E)-(H) represent out-of-domain datasets. To provide a comprehensive understanding of the model’s generalization capacity, we quantify the proportion of pre-training data associated with relevant anatomical structures and surgical procedures. These proportions are visualized using donut charts, with anatomical overlap depicted in red and procedural overlap shown in blue, effectively illustrating the degree of alignment between the pre-training data and downstream datasets. Evaluation metrics include image-level accuracy, video-level accuracy and phase-level Jaccard, providing a comprehensive assessment of model performance across multiple granularities. Statistical significance (P-value) is reported whenever SurgVISTA demonstrates superior performance compared to other methods. Detailed results are presented in Extended Data Tables~\ref{lab:Cholec80},~\ref{lab:M2CAI16-Workflow},~\ref{lab:PmLR50},~\ref{lab:AutoLaparo},~\ref{lab:ESD57},~\ref{lab:CATARACT},~\ref{lab:Cataract-101} and~\ref{lab:Cataract-21}.}
\label{natural_1}
\end{figure}

\subsection*{SurgVISTA outperforms natural-domain image- and video-level pre-trained methods on surgical tasks}

\begin{figure}[!ht]
\centering
\includegraphics[width=\linewidth]{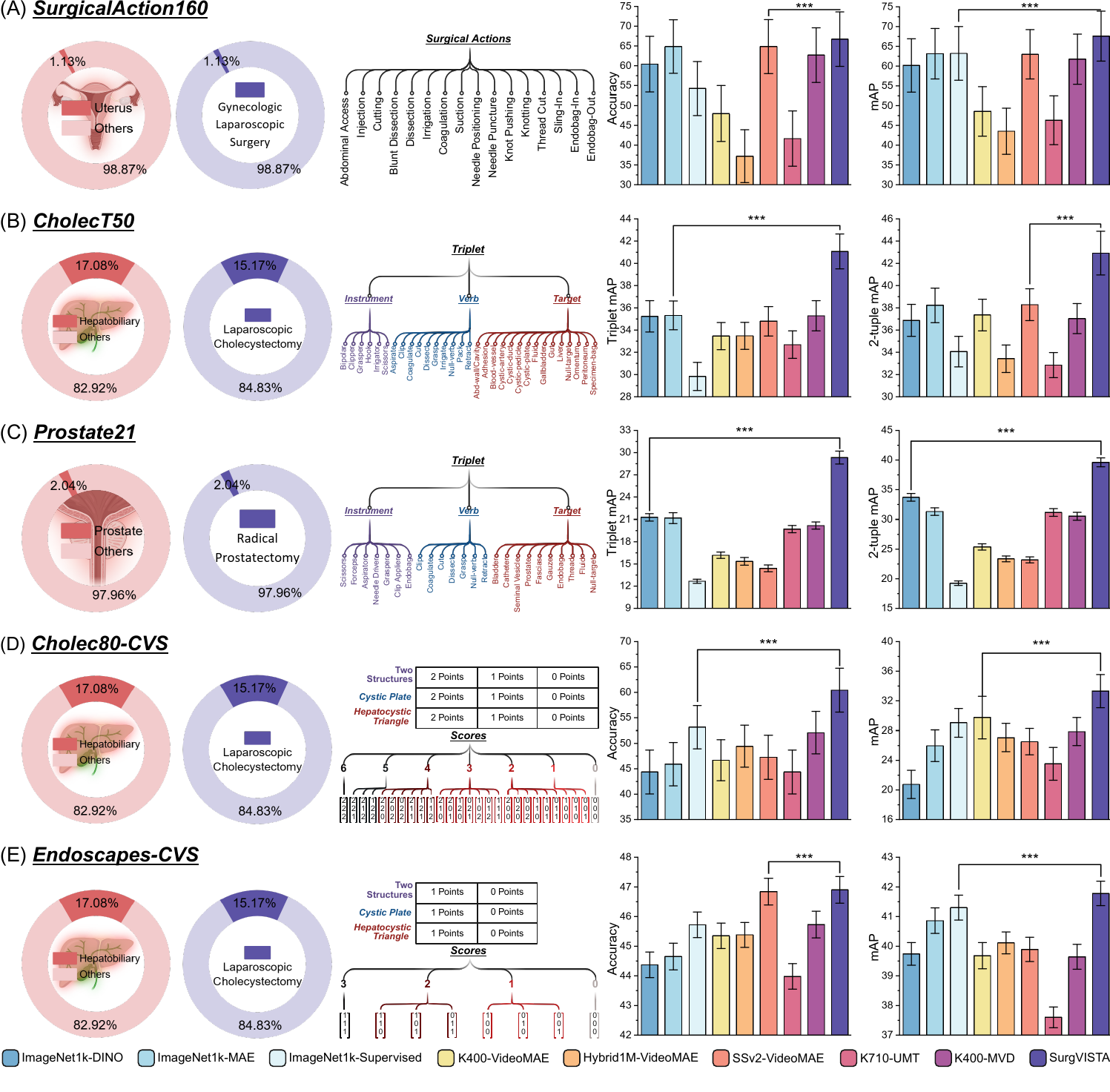}
\caption{Experimental results of various natural-domain pre-trained methods and SurgVISTA on surgical tasks beyond phase recognition. Subfigure (A) presents the results for the surgical action recognition task, with evaluation metrics reported as accuracy and mAP. Subfigures (B)-(C) illustrate the surgical triplet recognition task, with performance reported in terms of triplet mAP and 2-tuple mAP. Subfigures (D)-(E) display the results for the surgical skill assessment task, evaluated in terms of accuracy and mAP. Statistical significance (P-value) is reported whenever SurgVISTA demonstrates superior performance compared to other methods. Detailed results are presented in Extended Data Tables~\ref{lab:CholecT50},~\ref{lab:Prostate21}, and~\ref{lab:SurgicalActions160}.}
\label{natural_2}
\end{figure}

In this section, we first evaluate the effectiveness of SurgVISTA on surgical workflow recognition, comparing it against both image-level~\cite{ViT,MAE,DINO} and video-level~\cite{videomae,internvideo,MVD,UMT} pre-trained methods within the natural domain. Experimental results are illustrated in Fig.~\ref{natural_1}, accompanied by 95\% confidence intervals to quantify uncertainty and P-values to assess statistical significance. Specifically, Fig.~\ref{natural_1} (A)-(D) present results on in-domain datasets with related procedures in the pre-training data, while Fig.~\ref{natural_1} (E)-(H) illustrate results on out-of-domain datasets, involving surgical procedures excluded from the pre-training data. To provide a comprehensive understanding of the model’s generalization capacity, we quantify the proportion of pre-training data associated with related surgical procedures and anatomical structures, thereby highlighting the extent of overlap between the pre-training data and downstream datasets. As illustrated in Fig.~\ref{natural_1}, natural-domain methods employing video-level pre-training consistently outperformed image-level counterparts across all evaluated datasets. On Cholec80, the average gains for video-level models over image-level ones reached 1.5\%, 1.5\%, and 2.8\% in image-level accuracy, video-level accuracy, and phase-level Jaccard, respectively. This performance advantage became even more substantial on the out-of-domain Cataract-101 dataset, with average enhancements reaching 10.8\% in image-level accuracy, 10.7\% in video-level accuracy, and 19.6\% in phase-level Jaccard, highlighting the pronounced benefits of video-level pre-training for surgical scene understanding. Furthermore, SurgVISTA consistently outperformed natural-domain pre-trained video-level methods across both in-domain and out-of-domain datasets. On representative in-domain benchmarks including Cholec80, M2CAI16-Workflow, and PmLR50, SurgVISTA achieved notable gains in image-level accuracy, video-level accuracy, and phase-level Jaccard over natural-domain methods. Despite minimal procedural overlap in the pre-training data (e.g., 1.13\% for hysterectomy), SurgVISTA outperformed the best prior model~\cite{videomae} by 4.7\%, 4.6\%, and 7.3\%, highlighting its robust generalization to underrepresented procedures. These in-domain results demonstrate SurgVISTA’s capacity to acquire domain-relevant knowledge during pre-training, enabling improved downstream performance. In out-of-domain evaluations on ESD57, SurgVISTA surpassed VideoMAE~\cite{videomae} (pre-trained on UnlabeledHybrid~\cite{internvideo}) by 2.8\%, 2.7\%, and 5.4\% on the three metrics, respectively. When evaluated on CATARACTS, Cataract-101, and Cataract-21 datasets, which are substantially different from the surgical procedures used during pre-training, SurgVISTA consistently outperformed all competing methods, reinforcing its generalizability across varied procedures and anatomical structures. We further evaluate SurgVISTA on additional critical surgical tasks beyond workflow recognition, providing a comprehensive assessment of its capacity for surgical scene understanding. As illustrated in Fig.~\ref{natural_2}, SurgVISTA consistently achieved superior performance compared to natural-domain pre-trained methods across multiple surgical tasks. For surgical action recognition on SurgicalActions160, it surpassed UMT~\cite{UMT} by 1.8\% in accuracy and 4.6\% in mAP. For surgical triplet recognition, it outperformed MVD~\cite{MVD} on CholecT50 by 5.8\% (triplet mAP) and 5.9\% (2-tuple mAP), and exceeded DINO~\cite{DINO} on Prostate21 by 8.0\% and 5.9\%, suggesting effective modeling of instrument–target interactions. For surgical skill assessment, SurgVISTA achieved relative improvements of 8.3\% in accuracy and 5.4\% in mAP over MVD~\cite{MVD} on Cholec80-CVS, and surpassed UMT~\cite{UMT} with relative gains of 0.1\% in accuracy and 1.9\% in mAP on Endoscapes-CVS. These results highlight its capacity to capture intricate anatomical structures and motion patterns critical for evaluating surgical expertise.

Overall, these tasks constitute core components of comprehensive surgical scene understanding. Extensive evaluations across 13 diverse datasets demonstrate that SurgVISTA consistently surpasses natural-domain pre-trained methods across a wide range of tasks, surgical procedures, and anatomical structures.

\subsection*{SurgVISTA outperforms existing surgical-domain pre-trained methods}

\begin{figure}[!ht]
\vspace{5pt}
\centering
\includegraphics[width=\linewidth]{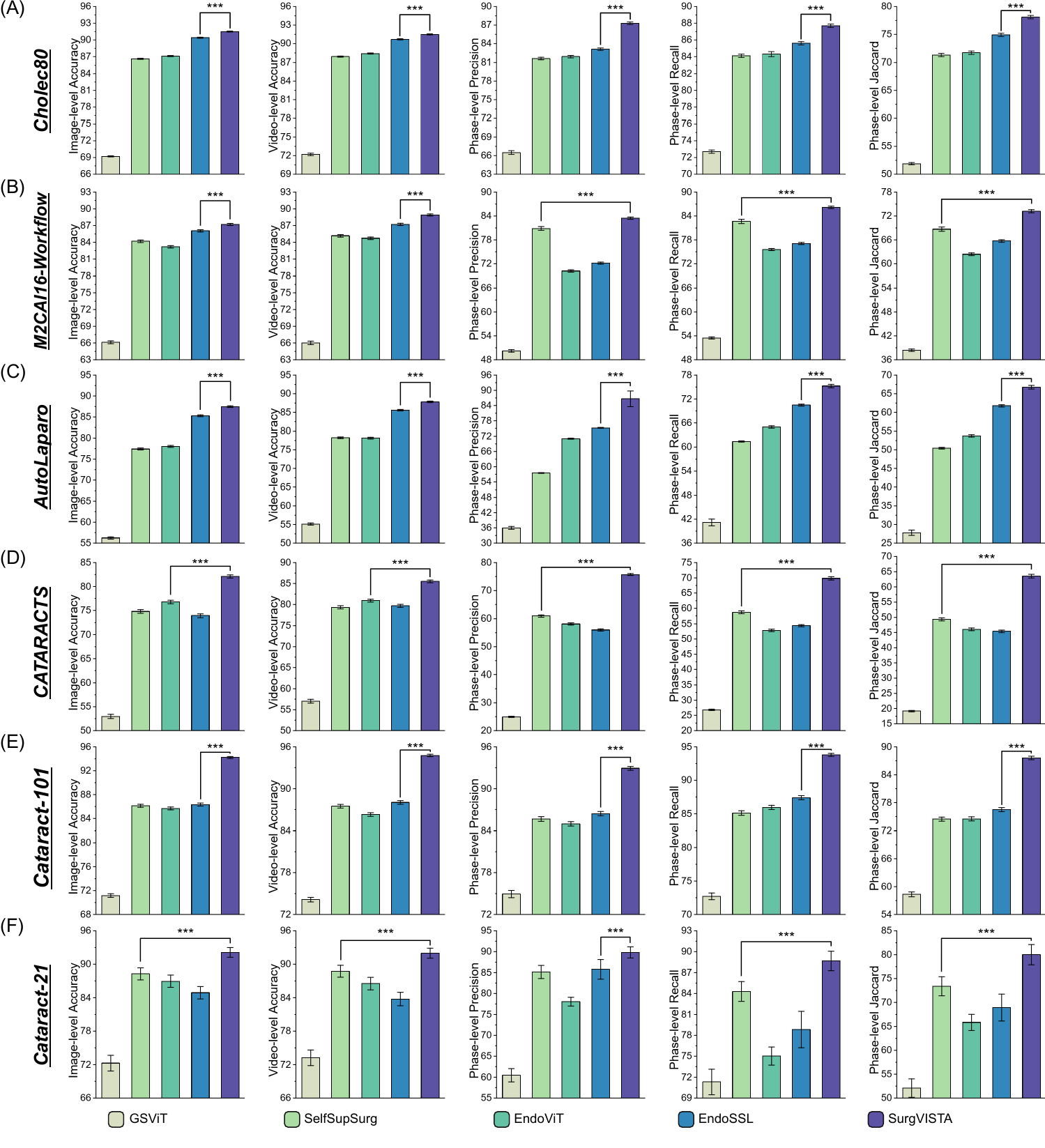}
\caption{
Experimental results comparing SurgVISTA and various surgical-domain pre-trained methods on surgical phase recognition datasets. Subfigures (A)-(C) correspond to in-domain datasets, while (D)-(F) represent out-of-domain datasets. Evaluation metrics include image-level accuracy, video-level accuracy, phase-level precision, phase-level recall and phase-level Jaccard, providing a comprehensive assessment across multiple granularities. Statistical significance (P-value) is reported whenever SurgVISTA  demonstrates superior performance compared to competing methods. Detailed results are presented in Extended Data
Tables~\ref{lab:Cholec80_surgical},~\ref{lab:M2CAI16-Workflow_surgical},~\ref{lab:AutoLaparo_surgical},~\ref{lab:CATARACTS_surgical},~\ref{lab:Cataract-101_surgical} and~\ref{lab:Cataract-21_surgical}.
}
\label{fig:surgical}
\end{figure}

To further investigate the effectiveness of SurgVISTA in surgical scene understanding, we conducted a comprehensive comparison with existing surgical-domain pre-trained methods. Given that current surgical-domain pre-trained methods~\cite{selfsupsurg,endovit,GSViT,endossl} primarily emphasize the surgical workflow recognition task, we performed extensive evaluations on six surgical workflow recognition datasets encompassing diverse procedures and anatomical structures. Experimental results with 95\% confidence intervals are presented in Fig.~\ref{fig:surgical}, providing a comprehensive comparison of model performance across evaluated datasets. Specifically, Fig.~\ref{fig:surgical} (A)-(C) represent results on in-domain datasets, while Fig.~\ref{fig:surgical} (D)-(E) illustrate the results on out-of-domain datasets. Overall, SurgVISTA significantly outperformed all comparative methods across all evaluation metrics and datasets. On in-domain datasets, SurgVISTA exhibited substantial improvements over the best-performing method EndoSSL~\cite{endossl}. Notably, EndoSSL is an image-level method pre-trained on 23.3 million private laparoscopic frames using the ViT-L architecture~\cite{ViT}. Specifically, on the Cholec80 dataset, SurgVISTA achieved high performance, with image-level accuracy of 91.5\%, video-level accuracy of 91.5\%, phase-level precision of 87.3\%, phase-level recall of 87.7\%, and phase-level Jaccard of 78.1\%, significantly outperforming other methods. On the M2CAI16-Workflow dataset, compared to EndoSSL~\cite{endossl}, SurgVISTA attained modest gains of 1.1\% and 1.7\% in image-level and video-level accuracies, while achieving significantly higher performance in phase-level metrics, with improvements of 11.3\% in precision, 9.1\% in recall, and 7.5\% in Jaccard. On the AutoLaparo dataset, SurgVISTA exhibited remarkable improvements across multiple granularities compared to EndoSSL~\cite{endossl}, where it achieved an increase of 2.2\% in image-level accuracy, 2.2\% in video-level accuracy, 11.4\% in phase-level precision, 4.8\% in phase-level recall, and 5.1\% in phase-level Jaccard. In out-of-domain evaluations involving surgical procedures not encountered during pre-training, SurgVISTA consistently achieved superior performance. For the CATARACTS dataset, containing numerous workflow categories and complex phase transitions, SurgVISTA surpassed all competing methods, and achieved improvements of 14.8\%, 11.1\% and 14.3\% in phase-level metrics compared to SelfSupSurg~\cite{selfsupsurg}. Similarly, for the Cataract-101 and Cataract-21 datasets, SurgVISTA consistently surpassed existing models.

\begin{figure}[!t]
\centering
\includegraphics[width=0.92\linewidth]{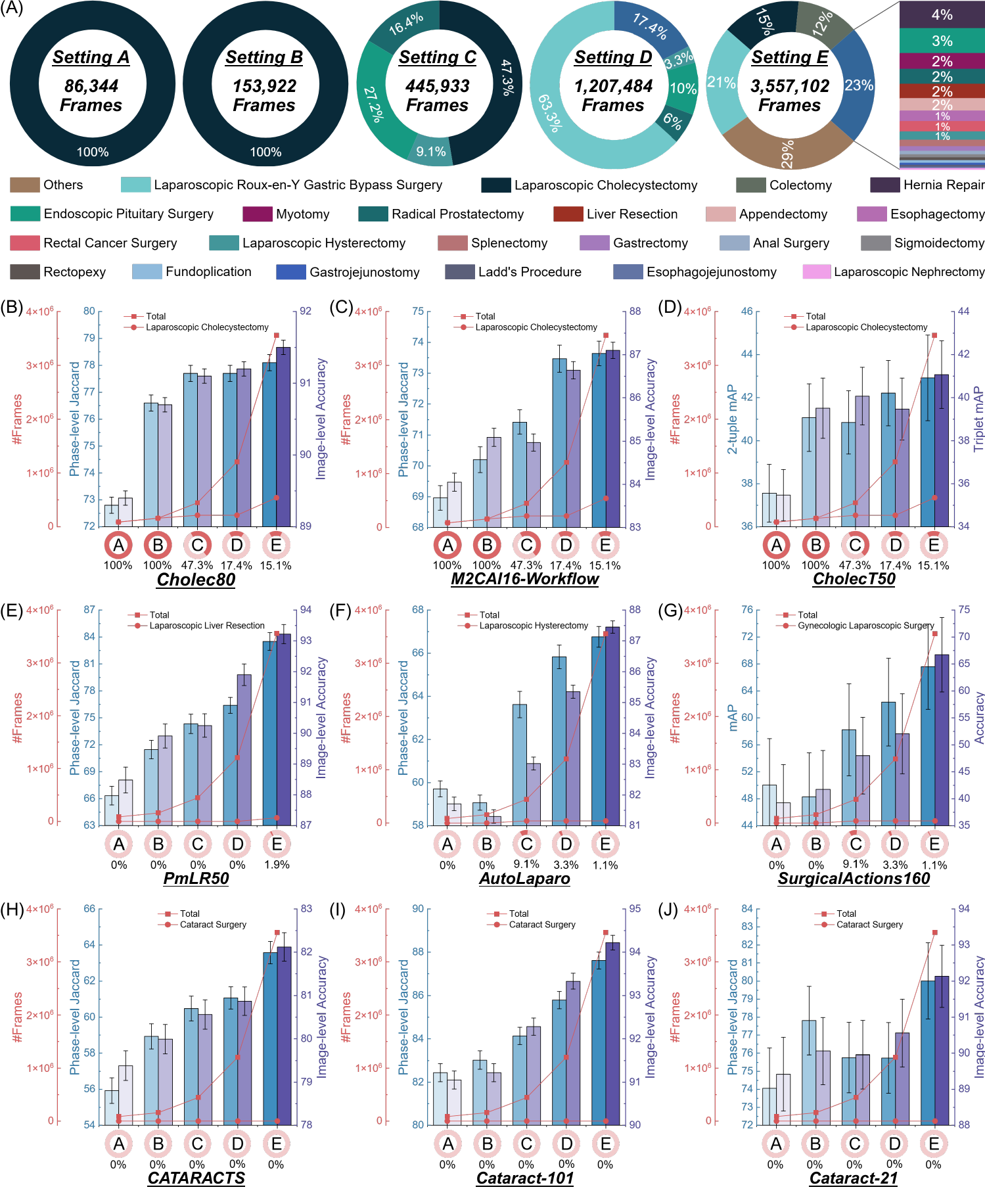}
\caption{Experimental overview of how varying pre-training data scales affect the performance and generalization. Subfigure (A) illustrates the distribution of surgical procedures across progressively constructed sub-datasets. Subfigures (B)-(J) present performance variations across nine surgical datasets under different data scales, along with data volumes and proportional distribution. Phase-level Jaccard and image-level accuracy are visualized using dual-axis bar plots: blue bars (left y-axis) denote phase-level Jaccard, while purple bars (right y-axis) represent image-level accuracy.
}
\label{fig:scaling}
\end{figure}
\begin{figure}[!t]
\centering
\includegraphics[width=\linewidth]{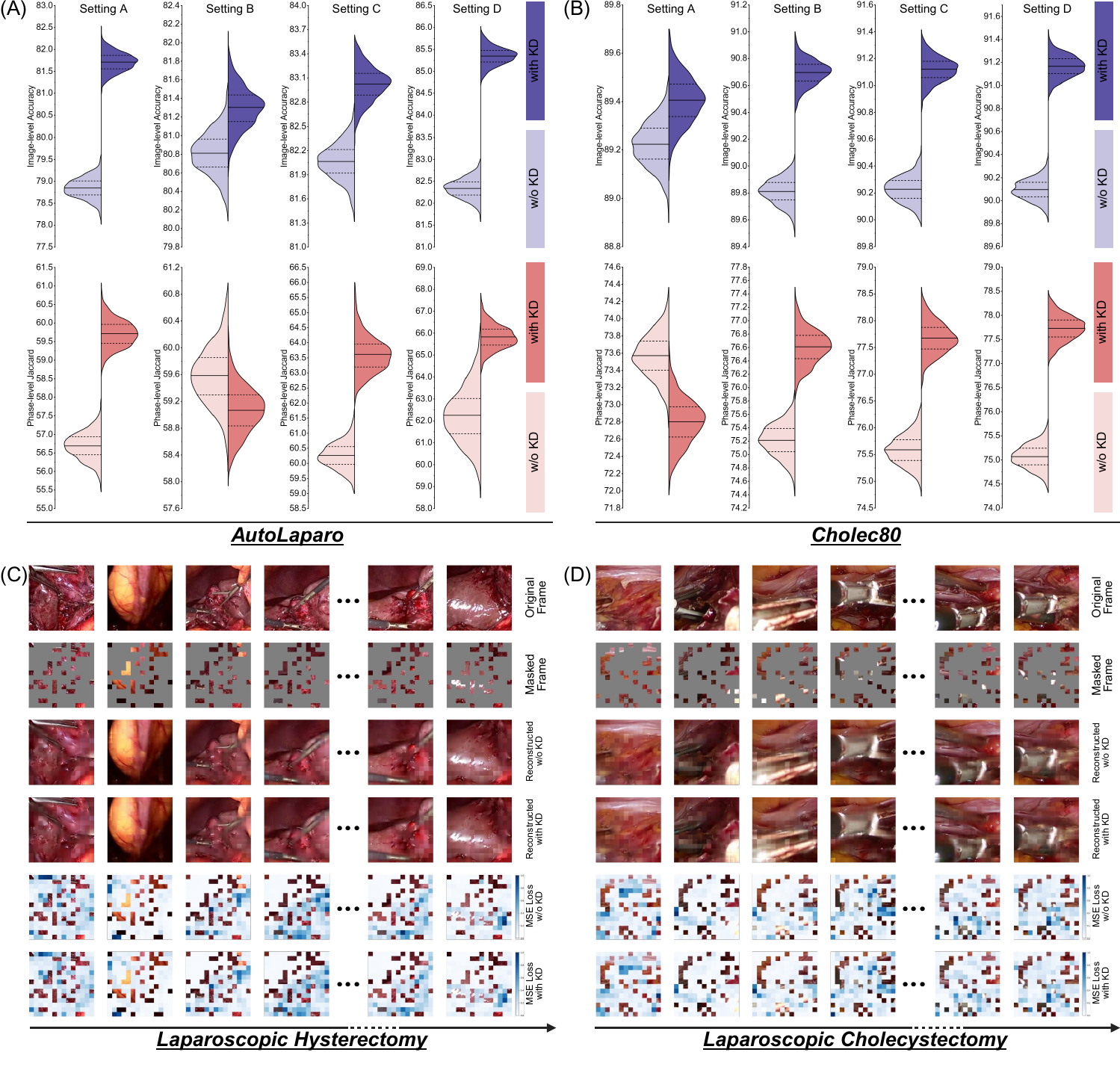}
\caption{The effectiveness of knowledge distillation. Subfigures (A)-(B) present the performance difference between SurgVISTA with and without knowledge distillation. Significance testing was conducted using the Wilcoxon signed-rank one-sided test, demonstrating that expert knowledge distillation consistently improves performance across different pre-trained sub-datasets and downstream datasets. Subfigures (C)-(D) qualitatively illustrate the reconstruction performance based on Setting D by showing the original frame, masked input, reconstructions with and without knowledge distillation, and corresponding MSE maps. The results demonstrate that knowledge distillation improves reconstruction fidelity by better preserving fine-grained anatomical details, as supported by reduced reconstruction error in related regions.}
\label{fig:kd}
\end{figure}
\subsection*{SurgVISTA benefits from large-scale and diverse surgical surgical pre-training data} 
Data scaling laws describe how model performance varies as the amount of pre-training data increases, serving as a fundamental principle in developing robust foundation models. Extensive natural-domain studies have demonstrated that scaling pre-training data significantly enhances model generalization, robustness, and overall performance across various tasks. However, the investigation of data scaling laws within the surgical domain, particularly for video-level pre-training, remains relatively underexplored. To elucidate the relationship between data scale and performance within the surgical domain, we systematically examined data scaling laws by pre-training SurgVISTA on a series of sub-datasets of varying sizes. As depicted in Fig.~\ref{fig:scaling} (A), we constructed five progressively expanding sub-datasets, denoted as Settings A through E, where each successive setting incorporated all data from the preceding settings, culminated in Setting E representing the full dataset. Sub-datasets from Settings A to D were exclusively sourced from SurgPub, with each subsequent setting incorporating additional public datasets to examine the impact of diverse surgical procedures on model performance and generalization capabilities. The detailed composition and distribution are illustrated in Extended Data Table~\ref{lab:pre-training_dataset}. To assess model performance and generalization capabilities, extensive experiments were conducted across nine diverse video-level surgical datasets, spanning various surgical procedures and anatomical structures. The experimental results are shown in Fig.~\ref{fig:scaling} (B)-(J), where line charts indicate the absolute data quantities of procedure-specific and full pre-training data, while the accompanying donut charts illustrate the corresponding proportions. Overall, as the scale of pre-training data progressively increases, all downstream datasets demonstrate a consistent and robust trend of performance enhancement. For Cholec80, M2CAI16-Workflow, and CholecT50, which focus on laparoscopic cholecystectomy procedures, we observed an overall trend of performance improvement from Setting A to Setting C, with the proportion of procedure-specific data gradually decreasing from 100\% to 47.3\%, indicating the model’s ability to effectively leverage relevant surgical content. Even in Settings D and E, where the proportion of procedure-specific data decreased to 17.4\% and 15.1\% respectively, consistent performance improvements were observed, highlighting the model’s robust generalization across low-overlap surgical scenarios. For PmLR50, which has 0\% procedural overlap with the pre-training data from Settings A to D, performance continued to improve, achieving gains of 3.4\% in image-level accuracy and 10.1\% in phase-level Jaccard. In Setting E, despite the procedure-specific data accounting for only 1.9\% of the total, the model achieved further improvements of 1.3\% in image-level accuracy and 7.1\% in phase-level Jaccard, demonstrating the effectiveness of large-scale surgical pre-training in enhancing representation transferability. For gynecologic laparoscopic surgery datasets, including AutoLaparo and SurgicalActions160, we observed a slight performance decline from Settings A to B, followed by a substantial improvement in Setting C, where the amount of pre-training data increases and procedure-relevant surgical videos were introduced. Specifically, from Setting A to Setting C, AutoLaparo demonstrated gains of 1.3\% in image-level accuracy and 3.9\% in phase-level Jaccard, while SurgicalActions160 showed improvements of 8.7\% in accuracy and 8.2\% in mAP. In subsequent settings, the continued inclusion of unrelated surgical procedures further enhanced performance, where AutoLaparo achieved an additional improvement of 4.5\% in image-level accuracy and 3.2\% in phase-level Jaccard, while SurgicalActions160 attained a further 18.7\% and 9.4\% increase in accuracy and mAP, respectively. For out-of-domain cataract surgery datasets, including CATARACTS, Cataract-21, and Cataract-101, we still observed significant performance improvements, demonstrating the strong generalization capability of large-scale surgical pre-training in leveraging diverse procedural contexts to learn robust and transferable spatiotemporal representations. Comprehensive analyses demonstrate that SurgVISTA exhibits substantial performance improvements as the scale of pre-training data expands, even when incorporating irrelevant surgical procedures, highlighting its robustness and generalization capacity. These performance improvements are particularly pronounced for datasets involving complex surgical procedures with numerous workflow categories and intricate phase transitions. The systematic investigation of data scaling laws in this study offers valuable insights into the relationship between data volume and model effectiveness, providing guidance for future research and the development of more advanced and generalizable models for surgical scene understanding.

\subsection*{Ablation studies confirm the necessity of knowledge distillation} 
Surgical scene understanding necessitates advanced spatiotemporal modeling to effectively capture the complex and dynamic interactions between anatomical structures and surgical instruments, wherein effective spatial modeling is essential for fine-grained intra-frame understanding and for enriching the overall spatiotemporal feature representation. To further improve spatial feature representations within the integrated spatiotemporal learning framework, SurgVISTA incorporates knowledge distillation, leveraging the superior spatial feature extraction capabilities of image-level foundation model. To assess the effectiveness of knowledge distillation, we implemented an ablated variant of SurgVISTA without knowledge distillation, pre-trained on Settings A through D, and conducted extensive experiments across two downstream datasets. The experimental results are illustrated in Fig.~\ref{fig:kd} (A)–(B). Under Setting A, knowledge distillation yielded marginal changes on Cholec80, with a slight increase of 0.2\% in image-level accuracy and a slight decrease of 0.6\% in phase-level Jaccard, possibly due to the inclusion of procedure-specific pre-training data aligned with Cholec80. In contrast, more substantial improvements were observed on AutoLaparo, with improvements of 2.9\% and 3.0\% in image-level accuracy and phase-level Jaccard, respectively. Furthermore, for Cholec80, the performance gains from knowledge distillation became increasingly evident as the scale and diversity of pre-training data increased. For AutoLaparo, although Settings A and B involved the same surgical procedure that differed from AutoLaparo’s, knowledge distillation yielded gains in Setting A but slightly decreased phase-level Jaccard in Setting B. As procedurally relevant data and overall dataset diversity increased in subsequent settings, the benefits of knowledge distillation became progressively more pronounced. Overall, knowledge distillation consistently improves performance, yielding significant improvements in both accuracy and Jaccard across two datasets and all four pre-trained settings, compared to the non-distilled counterparts. Complementary qualitative results are presented in Fig.~\ref{fig:kd} (C)-(D), where reconstruction fidelity is evaluated using mean squared error (MSE) to verify the model’s ability to preserve fine-grained visual details. These visualizations demonstrate that knowledge distillation facilitates the learning of robust spatial representations to reconstruct video clips with structural integrity and visual coherence, particularly in regions containing anatomical structures and instrument-tissue interactions.
\section{Discussion}\label{sec2}

To the best of our knowledge, this work is the first to explore video-level pre-training in the surgical domain using large-scale surgical video data. The following novel insights can be derived from our extensive experimental results.

\textbf{Surgery-specific video-level pre-training outperforms natural-domain counterparts by enabling domain-aligned spatiotemporal representation learning.}
Experimental results reveal two critical insights: temporal modeling capabilities acquired during pre-training are essential for capturing dynamic surgical contexts, and domain-aligned pre-training surpasses domain-agnostic pre-training in addressing complex and variable surgical scenarios. While natural-domain video pre-training surpasses image-level methods by better modeling temporal dynamics, it lacks the surgery-specific knowledge required for precise interpretation of instruments, anatomy, and their interactions. In contrast, surgical-domain video pre-training yields more robust and discriminative spatiotemporal representations enriched with surgery-specific knowledge, reinforcing its importance for robust surgical scene understanding.

\textbf{Video-level surgical pre-training outperforms image-level counterparts by enabling explicit spatiotemporal representation learning.} Although image-level counterparts primarily focus on spatial representation learning from large-scale surgical images, the absence of explicit temporal modeling during pre-training limits their ability to capture dynamic surgical contexts. Consequently, despite leveraging powerful surgery-specific spatial feature extractors, these methods inherently struggle to model temporal dependencies from purely spatial representations during fine-tuning, ultimately limiting spatiotemporal representation quality, particularly in complex surgical scenarios. SurgVISTA employs spatiotemporal representation learning during pre-training, effectively harnessing the intrinsic spatial and temporal cues embedded in video data. By integrating fine-grained spatial representations of instruments and anatomical structures with comprehensive temporal modeling encompassing action temporality, causality, and dynamic interactions, SurgVISTA provides a robust and generalizable foundation for surgical scene understanding. The demonstrated superiority highlights the critical necessity of video-level pre-training tailored to the surgical domain, facilitating the advancement of more sophisticated, reliable, and clinically applicable intelligent surgical systems.

\textbf{Increasing the scale and diversity of surgical video data yields substantial improvements in both performance and generalization.} Extensive experiments conclusively demonstrate that both factors are pivotal for improving accuracy and robustness across diverse surgical scenarios. First, scaling up pre-training data consistently enhances model performance and generalization, as reflected by SurgVISTA’s significant gains across all evaluated datasets. In procedure-agnostic scenarios involving procedures not present in downstream datasets, SurgVISTA continues to benefit by learning more diverse visual and temporal patterns. In procedure-specific scenarios, incorporating additional relevant data reinforces domain-specific priors and facilitates discriminative feature learning, further improving task-specific performance. This highlights the importance of large-scale surgical video data for advancing comprehensive understanding of complex and diverse surgical scenarios. Second, increasing procedural diversity further boosts performance and generalization. As more diverse surgical procedures are incrementally integrated into the pre-training corpus, significant performance gains are observed across various surgical scenarios, indicating that learning from a wider procedural spectrum enables the model to capture shared spatial and temporal patterns that generalize beyond specific tasks.  Enhancing procedural diversity facilitates the exploration of universal and transferable representations essential for understanding complex and diverse surgical scenarios.

\textbf{Data splitting in surgical video analysis remains fundamentally constrained by the trade-off between maximizing training scale and preserving unbiased evaluation.} Given the inherent complexity and procedural overlap in surgical video data, including shared instruments, anatomical structures, or activities, expanding the training set inevitably increases the risk of distributional leakage into the evaluation data. This presents a fundamental challenge: how to maximize training scale for generalization while rigorously maintaining evaluation integrity. In this study, we curated publicly available surgical video datasets with any form of annotation. To enhance the diversity and scale of pre-training, we selected several datasets specifically for representation learning. Among these, some followed their official training/testing splits to support downstream evaluation under strict isolation, while others were used exclusively during pre-training to expand data scale. Additionally, a number of datasets were entirely excluded from the pre-training phase and reserved as unseen evaluation benchmarks to assess generalization across surgical domains. This strategy enables a rigorous assessment of model generalization while ensuring strict non-overlap between pre-training and evaluation sets. Although dynamic benchmarks like LiveBench have emerged in general machine learning, such adaptive evaluation remains challenging in surgical domains due to limited data access, high annotation costs, and strict privacy constraints.

\textbf{SurgVISTA represents a robust and generalizable surgical video foundation model with clear advantages over both natural-domain and surgery-specific methods.} Compared to natural-domain methods, SurgVISTA leverages large-scale surgical video data to acquire domain-specific knowledge that encompasses both static anatomical structures and surgical instruments, as well as dynamic intraoperative activities. This domain alignment results in more clinically relevant and effective performance in intelligent surgical applications. In contrast to existing surgery-specific methods, SurgVISTA pioneers a unified spatiotemporal representation learning paradigm, marking a transition from static image-based representation learning to dynamic video-level understanding. Through explicit modeling of spatiotemporal representation inherent in large-scale surgical video data, SurgVISTA demonstrates superior capacity to handle the complexity and variability of real-world surgical scenarios and exhibits improved generalization and transferability across heterogeneous procedural contexts. By delivering transferable and discriminative spatiotemporal features and enabling seamless adaptation to a broad range of downstream tasks, SurgVISTA serves as a solid foundation for the development of intelligent surgical systems with the potential to improve procedural safety and enhance clinical outcomes across diverse surgical scenarios. These results affirm that the generalized and robust spatiotemporal representations learned by SurgVISTA from large-scale surgical video data are pivotal for accurately interpreting complex surgical scenarios, significantly outperforming natural-domain methods lacking specialized surgical knowledge. 

\textbf{Limitations and future directions.} While SurgVISTA represents a substantial advancement in surgical video pre-training and spatiotemporal representation learning, several limitations remain, highlighting promising directions for future investigation. First, although we construct a large-scale video-level surgical dataset encompassing more than 20 surgical procedures and over 10 anatomical structures, it remains insufficient to fully reflect the extensive procedural complexity and variability in real-world clinical practices. This limitation underscores the imperative for continuous expansion and diversification of the pre-training dataset to ensure broader coverage of surgical scenarios and enhance the generalizability of learned representations across diverse clinical environments. Beyond sheer data scale, future work should also explore data-centric research questions in surgical SSL. In particular, training foundation models under fixed dataset sizes with varying procedure distributions offers an intriguing direction for understanding how data composition affects model generalization. Second, while we establish the most comprehensive video-level benchmark encompassing 13 downstream datasets spanning six surgical procedures and four distinct tasks, the procedural diversity within the benchmark remains limited relative to the broader scope of the pre-training data. Future efforts should focus on extensive collaboration with clinical institutions to collect annotated datasets across various surgical procedures, focusing on the generation of downstream tasks of high clinical relevance. Finally, although SurgVISTA achieves superior performance across diverse benchmark tasks, thorough clinical validation is essential to determine its practical utility and impact in real-world surgical scenarios. Effective deployment of SurgVISTA-driven systems necessitates both the development of efficient and task-aligned downstream functionalities and close collaboration with clinical experts to ensure seamless integration into surgical workflows and alignment with clinical standards and practical requirements.
\section{Method} \label{sec4}

\subsection*{Architecture Design}
SurgVISTA is the first video-level self-supervision paradigm specifically targeted to surgical video analysis. The core innovation is a dual-decoder design designed for simultaneous holistic video understanding and fine-detail preservation. As illustrated in Fig.~\ref{overview} (B), a video-level reconstruction branch enables the learning of motion dynamics and global contextual features while an image-level distillation branch ensures the retention of fine-grained spatial semantics and structural details. Specifically, given a video clip $X_v \in \mathbb{R}^{T\times H\times W\times 3}$, we employ a joint space-time cube embedding strategy~\cite{videomae} with a cube size of $2\times16\times16$ to partition the clip into multiple non-overlapping patches arranged in spatiotemporal order. Each patch is mapped to a visual token, yielding spatiotemporal representations $F_v\in\mathbb{R}^{T\times K\times C}$. To enhance the effectiveness of masked video modeling and reduce temporal redundancy, we apply tube masking with a masking ratio to 85\% to minimize information leakage while facilitating spatiotemporal representation learning. The encoder employs the vanilla ViT-B~\cite{ViT} architecture, augmented with a joint spatiotemporal attention mechanism to effectively capture inter-frame dynamics and intra-frame spatial structures. Following the encoder, video-level reconstruction focuses on restoring low-level pixel information throughout the entire video clip, thereby facilitating the learning of holistic spatiotemporal representations. The enhanced visible tokens from the encoder are concatenated with learnable mask tokens and then fed into a lightweight spatiotemporal decoder, which infers the missing pixel-level details with structural and temporal consistency from partially visible inputs. The reconstruction loss is defined as the mean squared error (MSE) between the normalized ground-truth pixel values and their reconstructed counterparts, computed exclusively within the masked regions. By reconstructing masked portions from visible token representations, SurgVISTA is guided to attend to the temporal dynamics within surgical contexts, thereby facilitating the extraction of semantically meaningful spatiotemporal features. Furthermore, while the cube embedding strategy enables efficient temporal down-sampling and mitigates the computational overhead of video-level pre-training, it inevitably compromises fine-grained spatial details essential for comprehensive spatiotemporal interpretation. To address this limitation, image-level knowledge distillation is employed to optimize high-level spatial representations by leveraging supervision from a surgery-specific expert model~\cite{endossl}. Specifically, the encoder’s enhanced visible tokens are concatenated with learnable mask tokens and fed into a lightweight spatiotemporal decoder to generate enhanced spatial representations, which are subsequently aligned with the frame-level spatial features extracted from the expert model. The distillation loss is defined as the Smooth L1 loss between the predicted features and the  expert-derived spatial representations.

\subsection*{Data Splitting Strategy}

Data leakage poses a significant challenge in the benchmarking of foundation models~\cite{white2024livebench}. To build a comprehensive and diverse corpus for both development and evaluation, we systematically collected a wide range of publicly available surgical video datasets. To avoid any contamination between the pre-training and evaluation phases, we strictly followed the principles below. First, no videos used during the pre-training phase were included in the test splits of any downstream datasets. For several datasets utilized in both pre-training and evaluation, all test samples were explicitly excluded from the pre-training data to prevent data leakage. Second, we conducted in-domain evaluations using two types of datasets: (1) the held-out test splits of benchmark datasets partially used in pre-training, and (2) additional datasets involving surgical procedures that are consistent with those seen during pre-training, but whose video samples were not included in the pre-training phase. These evaluations assessed the model’s ability to generalize within familiar clinical domains while avoiding potential data leakage. Third, to assess generalization under distribution shifts, we conducted out-of-domain evaluations using datasets whose surgical procedures were entirely absent from the pre-training corpus. Overall, these precautions ensured that no test split from any downstream evaluation contained videos seen during pre-training, allowing both in-domain and out-of-domain generalization to be assessed under rigorously controlled conditions.

\subsection*{Implementation Details}

For \textbf{pre-training}, we initialized SurgVISTA entirely from scratch, employing random initialization for all components. The encoder adopted the ViT-B architecture with fixed spatiotemporal positional embeddings and omitted the use of a class token. The decoder designed for video-level reconstruction comprised four conventional transformer layers, while the decoder dedicated to image-level knowledge distillation employed a two-layer transformer architecture. The total number of parameters was 100.8 M, with the encoder accounting for 87.4 M. The pre-training dataset consisted of 3,650 videos, totaling approximately 3.55 million frames, spanning diverse surgical procedures across distinct anatomical structures. For each video, we first sampled frames at 1 frame per second (fps), and subsequently constructed input video clips by uniformly sampling 16 frames at a fixed interval of 4, starting sequentially from each frame. The model was pre-trained for 200 epochs with a batch size of 512, employing gradient accumulation over four iterations. We employed the AdamW optimizer with $\beta_1 = 0.9$, $\beta_2 = 0.95$, an initial learning rate of 1.5e-4, and a weight decay of 0.05. These hyperparameters were directly adopted from prior work~\cite{videomae} without additional tuning, except for the number of epochs, which was proportionally adjusted according to the size of the surgical video dataset to ensure sufficient convergence. To balance the dual objectives of spatiotemporal modeling and spatial representation refinement, the video-level reconstruction loss was assigned a weight of 1.0, while the image-level knowledge distillation loss was assigned a weight of 0.05. The pre-training of SurgVISTA was performed on 8 NVIDIA H800 GPUs for 10 days. Following standard practice in self-supervised learning~\cite{videomae,MVD}, no downstream validation data were used to guide the pre-training process. The final model checkpoint was selected at the end of training, at which point the model had already converged. For \textbf{fine-tuning}, we retained the architectural configurations of each pre-trained method to ensure faithful reproduction and fair comparison. For natural-domain methods, the encoder was initialized with the corresponding pre-trained parameters, and a single-layer fully connected (FC) head was appended for prediction, with all task-specific layers initialized randomly. The primary distinction between image-level and video-level methods lay in the embedding mechanism: image-level models employed standard 2D convolutional embeddings, while video-level models adopted cube embeddings via 3D convolutions. All training configurations were kept consistent across SurgVISTA and natural-domain methods on each dataset, except for the input frame length, which was set to 16 frames for video-level methods and 8 frames for image-level ones, to maintain comparable computational overhead. For comparisons with surgical-domain methods, SurgVISTA retained its end-to-end framework, consisting of an encoder followed by a single-layer FC head. In contrast, other surgical-domain methods were implemented following the conventional two-stage paradigm~\cite{TeCNO}, where the encoder was first fine-tuned using frame-level supervision, followed by a temporal convolutional module that aggregated the extracted spatial features for final prediction. To ensure training consistency and fair comparison, we followed the official training configurations provided in the original implementation~\cite{TeCNO}. For rigorous \textbf{downstream evaluation}, we adopted a strict data partitioning protocol as summarized in Extended Data Table~\ref{lab:pre-training_dataset}. For all in-domain downstream datasets, dedicated test sets were explicitly excluded from both the pre-training and fine-tuning phases. For out-of-domain evaluations, the entire dataset, including both training and testing splits, was completely disjoint from the pre-training data. This protocol ensured a fair assessment of both domain-specific performance and cross-domain generalization.

\subsection*{Evaluation Benchmark}

To comprehensively assess the performance of SurgVISTA, we curated a video-level evaluation benchmark encompassing six distinct surgical procedures across five anatomical structures, including laparoscopic cholecystectomy, laparoscopic hysterectomy (gynecologic laparoscopic surgery), laparoscopic liver resection, endoscopic pituitary surgery, ophthalmic surgery, and endoscopic submucosal dissection. This benchmark contained 13 datasets, comprising more than 500 long-form videos and 600 short-form clips, totaling over 660,000 frames. Detailed statistical information of these downstream datasets was provided in Extended Data Table~\ref{downstream}. To prevent potential data leakage, we strictly ensured that the test sets from in-domain downstream datasets overlapping with the SurgPub were entirely excluded from both the pre-training and fine-tuning phases. Furthermore, we conducted experiments across four clinically significant tasks, including surgical workflow recognition, surgical action recognition, surgical triplet recognition, and surgical skill assessment. Effective performance on these tasks is essential for comprehensive surgical scene understanding, facilitating more precise comprehension of complex and dynamic surgical scenarios.

\subsubsection*{Surgical Workflow Recognition}
Surgical workflow recognition serves as a fundamental component of surgical scene understanding, which aims to automatically assign predefined surgical phases to individual frames within surgical video. In clinical practice, surgical procedures are systematically partitioned into well-defined phases based on expert knowledge, with each phase representing a specific objective or set of actions performed during the operation. Precise phase recognition is crucial for understanding procedural context and tracking intraoperative progress, thereby enabling real-time surgical assistance, procedural monitoring, and intraoperative decision support. To comprehensively evaluate the effectiveness of our approach on surgical workflow recognition, we conducted extensive experiments across seven public datasets: Cholec80~\cite{endonet}, M2CAI16-Workflow~\cite{endonet,m2cai16workflow}, PmLR50~\cite{PmLR50}, AutoLaparo~\cite{AutoLaparo}, CATARACTS~\cite{Cataracts}, Cataract-101~\cite{cataract-101}, and Cataract-21~\cite{cataract21}, as well as one private dataset, ESD57. These datasets collectively covered a wide range of surgical procedures and anatomical structures, enabling a comprehensive assessment of the model’s generalizability across diverse clinical scenarios. Following standard practice in surgical workflow recognition~\cite{endonet,m2cai16workflow,TMRNet,yang2024surgformer,yang2024surgpetl}, performance was evaluated across multiple granularities using five widely adopted metrics: image-level accuracy, video-level accuracy, phase-level precision, phase-level recall, and phase-level Jaccard. These metrics capture performance at three levels of granularity: image-level accuracy treats each frame independently to assess global frame-wise correctness; video-level accuracy measures average performance across individual videos, reflecting video-wise correctness; and phase-level metrics assess category-wise performance across the dataset, reflecting the model’s ability to distinguish and localize surgical phases.

\subsubsection*{Surgical Action Recognition}
Surgical action recognition seeks to identify intraoperative actions at the frame or clip level, facilitating detailed analysis of surgical activities. In contrast to coarse-grained workflow recognition, it focuses on the temporal and semantic interpretation of surgical maneuvers, such as dissecting, retracting, or suturing, which are critical determinants of patient safety and postoperative outcomes. In clinical practice, accurate action recognition plays a pivotal role in enhancing intraoperative situational awareness, enabling context-specific assistance such as real-time procedural guidance, anticipatory error prevention, and adaptive decision-making. By providing fine-grained semantic interpretation of intraoperative activities, this task lays the foundation for improving surgical precision, reducing complication rates, and ensuring procedural safety. To evaluate our approach on surgical action recognition, we conducted experiments on the SurgicalActions160~\cite{surgicalactions160} dataset, which encompasses a diverse set of intraoperative actions involving various surgical instruments and anatomical targets, thereby enabling a comprehensive assessment of model performance across varied surgical activity patterns. For surgical action recognition, performance was evaluated using accuracy, which measures overall classification correctness, and mean Average Precision (mAP), which offers a class-aware assessment by integrating precision-recall curves across all action categories.

\subsubsection*{Surgical Triplet Recognition}
Surgical triplet recognition focuses on fine-grained activity analysis by modeling detailed interactions among surgical instruments, actions (verbs), and anatomical targets. This task involves recognizing individual components and systematically modeling their interrelationships to capture intricate instrument-tissue interactions, thereby enhancing real-time surgical assistance, procedural monitoring, and informed intraoperative decision support. To assess the effectiveness of our approach on surgical triplet recognition, we conducted experiments on one public dataset, CholecT50~\cite{CholecT50}, and one private dataset, Prostate21. Both datasets contained annotated surgical videos with detailed triplet labels comprising $<$instrument, action, target$>$, which served as benchmarks for evaluating the model’s ability to understand instrument–tissue interactions. Consistent with prior work in surgical triplet recognition~\cite{CholecT50,tripnet}, we reported component average precision and triplet average precision. Component average precision evaluates the individual elements of the triplet, including instrument (AP\textsubscript{I}), verb (AP\textsubscript{V}), and target (AP\textsubscript{T}), based on the area under the precision-recall curve for each class. Triplet average precision measures the model’s ability to recognize interactions by considering combinations such as instrument-verb (AP\textsubscript{IV}), instrument-target (AP\textsubscript{IT}), and instrument-verb-target (AP\textsubscript{IVT}). The primary metric in this study is AP\textsubscript{IVT}, which captures the accuracy of full triplet recognition.

\subsubsection*{Surgical Skill Assessment}
Surgical skill assessment aims to evaluate specific actions or activities at the frame or clip level, requiring a comprehensive understanding of complex anatomical structures and spatiotemporal relationships. Accurate skill assessment is crucial for reducing the risk of injury to critical anatomical structures and promoting procedural standardization, thereby facilitating intraoperative monitoring and improving postoperative evaluation. To comprehensively evaluate the effectiveness of our approach in surgical skill assessment, we conducted evaluations on the Cholec80-CVS~\cite{Cholec80-CVS} and Endoscapes-CVS~\cite{Endoscapes-CVS201} datasets. Both datasets focused on assessing Strasberg’s Critical View of Safety (CVS) criteria, serving as a benchmark for evaluating surgical proficiency and promoting procedural safety. For surgical skill assessment, we adopted accuracy and mean Average Precision (mAP) as evaluation metrics, where accuracy reflects overall prediction correctness, and mAP provides a class-wise evaluation of recognition performance.

\subsection*{Experimental settings}

\subsubsection*{State-of-the-art pre-trained methods in the natural domain}
To enable a comprehensive and structured comparison, we evaluate eight representative natural-domain pre-trained methods, categorized into two groups: \textbf{image-based pre-training} and \textbf{video-based pre-training}. All models are built upon the ViT-B architecture~\cite{ViT} and are pre-trained on large-scale general-purpose datasets, without any additional fine-tuning on task-specific or dataset-specific natural-domain benchmarks. Detailed configurations are provided in Extended Data Table~\ref{tab:natural_params}. 

\textbf{Image-based Pre-trained Methods.} These approaches exclusively focus on learning spatial representations and are pre-trained using large-scale static image datasets. 

\begin{itemize}
\item
\textbf{ImageNet-1K}~\cite{imagenet} (\textbf{Supervised}~\cite{ViT}) refers to the pre-trained parameters obtained through fully supervised training on the ImageNet-1K dataset, which contains over 1.2 million labeled images across 1,000 object categories. 
\item
\textbf{ImageNet-1K}~\cite{imagenet} (\textbf{DINO}~\cite{DINO}) refers to the pre-trained parameters obtained by applying the DINO framework to the ImageNet-1K dataset, which learns semantic feature representations through self-distilled contrastive learning with momentum encoders and clustering.
\item
\textbf{ImageNet-1K}~\cite{imagenet} (\textbf{MAE}~\cite{MAE}) refers to the pre-trained parameters derived by applying the MAE framework to the ImageNet-1K dataset, which learns high-quality spatial representations by reconstructing pixel-level information from randomly masked image patches.
\end{itemize}

\textbf{Video-based Pre-trained Methods.} These methods are trained directly on large-scale video data to learn joint spatiotemporal representations by capturing both dynamic temporal patterns and spatial semantics.
\begin{itemize}
\item 
\textbf{Kinetics-400}~\cite{k400} (\textbf{VideoMAE}~\cite{videomae}) refers to the pre-trained parameters obtained by applying the VideoMAE framework to the Kinetics-400 dataset, a large-scale human action dataset containing approximately 240,000 video clips across 400 action classes. VideoMAE leverages a masked autoencoder architecture tailored for video inputs, enabling the model to learn robust spatiotemporal representations by reconstructing masked video patches.
\item 
\textbf{UnlabeledHybrid}~\cite{internvideo} (\textbf{VideoMAE}~\cite{videomae}) refers to the pre-trained parameters obtained by applying the VideoMAE framework to the UnlabeledHybrid dataset, which aggregates over 1.37 million video clips from Kinetics-710~\cite{k710}, Something-Something V2~\cite{ssv2}, AVA~\cite{ava}, WebVid2M~\cite{WebVid2M}, and other online sources. 
\item 
\textbf{Something-Something V2}~\cite{ssv2} (\textbf{VideoMAE}~\cite{videomae}) refers to the pre-trained parameters obtained by applying the VideoMAE framework to the Something-Something V2 (SSV2) dataset, which contains approximately 169,000 video clips across 174 action categories with object-centric temporal interactions.
\item 
\textbf{Kinetics-710}~\cite{k710} (\textbf{UMT}~\cite{UMT}) refers to the pre-trained parameters obtained by applying the UMT framework to the Kinetics-710 dataset, which merges Kinetics-400~\cite{k400}, Kinetics-600~\cite{k600}, and Kinetics-700~\cite{k700} while removing duplicates. Specifically, we adopt parameters from the first training stage of UMT, where masked video modeling is guided by CLIP-based supervision~\cite{CLIP} in a single-modality setting.

\item 
\textbf{Kinetics-400}~\cite{k400} (\textbf{MVD}~\cite{MVD}) refers to the pre-trained parameters obtained by applying the MVD framework to the Kinetics-400 dataset. MVD employs masked feature modeling combined with a teacher-student distillation framework, utilizing both image-level and video-level teacher features to effectively guide the learning of robust and coherent spatiotemporal representations.
\end{itemize}

\subsubsection*{State-of-the-art pre-trained methods in the surgical domain}

In this study, we select four state-of-the-art self-supervised surgical models as surgical foundation models to facilitate a comprehensive and complementary comparison. Detailed information is provided in Extended Data Table~\ref{tab:surgical_params}. 
\begin{itemize}
\item
\textbf{SelfSupSurg}~\cite{selfsupsurg} explores the application of four advanced image-level SSL methods, namely MoCo v2~\cite{MoCo_v2}, SimCLR~\cite{SimCLR}, DINO~\cite{DINO}, and SwAV~\cite{SwAV}, for surgical scene comprehension. For comparison, we adopt the DINO-based variant, which employs a ResNet-50~\cite{resnet} backbone pre-trained on the Cholec80~\cite{endonet} dataset containing a limited amount of surgical data. The study provides a performance evaluation across six datasets representing two distinct surgical procedures.
\item 
\textbf{EndoViT}~\cite{endovit} compiles a large-scale publicly available image-level endoscopic dataset, Endo700k, sourced from nine public datasets and consisting of over 700,000 images. It adopts the MAE~\cite{MAE} framework with a ViT-B backbone for domain-specific pre-training on the Endo700k dataset. The study evaluates model performance on three downstream datasets, all corresponding to the same surgical procedure.
\item 
\textbf{EndoSSL}~\cite{endossl} constructs an extensive private dataset consisting of 7,877 laparoscopic procedure videos, encompassing 23.3 million frames and spanning over five distinct surgical procedures. It employs the MSN~\cite{MSN} framework with a ViT-L backbone pre-trained on this private dataset, resulting in a strong image-level surgical foundation model. The study assesses model performance exclusively on the Cholec80 dataset, focusing on surgical workflow recognition task.
\item 
\textbf{GSViT}~\cite{GSViT} develops an efficient model that employs EfficientNet\cite{efficientvit} as the backbone and utilizes next-frame prediction as the pretext task for pre-training, providing weak temporal modeling capabilities. To facilitate this, it open-sources a large-scale surgical video dataset comprising over 680 hours of surgical footage sourced from websites, totaling more than 2.5 million frames and spanning 28 distinct surgical procedures. The evaluation of model performance is conducted solely on the Cholec80 dataset, with a specific focus on surgical workflow recognition task.
\end{itemize}

\subsubsection*{Pre-training Datasets}
In this study, we curate eight publicly available video-level surgical datasets, referred to as SurgPub, along with various surgical videos sourced from online platforms, designated as SurgWeb. Detailed information is provided in Extended Data Table~\ref{lab:pre-training_dataset}. 

\begin{itemize}
\item 
\textbf{Cholec80}~\cite{endonet} is a widely used laparoscopic surgery dataset focusing on the cholecystectomy procedure, comprising 80 videos in total. The dataset is officially divided into training and test sets, each consisting of 40 videos. For pre-training, we exclusively utilize the 40 videos from the training set, amounting to a total of 86,344 frames.

\item 
\textbf{MI2CAI16-Workflow}~\cite{endonet,m2cai16workflow} is a dataset curated for surgical workflow recognition, containing 41 laparoscopic videos of cholecystectomy procedures. The dataset is split into 27 videos for training and 14 videos for testing. We incorporate the 27 videos from the training split into the pre-training dataset, contributing a total of 67,578 frames.

\item 
\textbf{HeiChole}~\cite{HeiChole} is an open benchmark for laparoscopic cholecystectomy, focusing on surgical workflow and skill analysis. The dataset comprises 24 videos allocated for training, while the remaining 9 videos are designated for testing. For pre-training, we utilize the publicly available 24 videos from the training set, providing 55,139 frames in total.
\item 
\textbf{PitVis}~\cite{PitVis} is a dataset curated for step and instrument recognition in endoscopic pituitary surgery videos. It contains 25 publicly accessible videos comprising 120,018 frames. The entire dataset is incorporated into the pre-training phase to enhance the procedural diversity of the pre-training data.
\item 
\textbf{PSI-AVA}~\cite{PSI-AVA} is a dataset focused on robot-assisted radical prostatectomy procedures, designed to enhance holistic surgical scene understanding. It encompasses approximately 20.45 hours of surgical footage, comprising 8 videos with a total of 72,318 frames. The entire dataset is incorporated into the pre-training process.
\item 
\textbf{AutoLaparo}~\cite{AutoLaparo} is a widely used laparoscopic surgery dataset focusing on the hysterectomy procedure. It contains 21 videos, with 10 videos allocated for training and the remaining 11 designated for testing. We incorporate the 10 videos from the training set into the pre-training dataset, contributing a total of 40,211 frames.
\item 
\textbf{BernBypass70}~\cite{MultiBypass140} consists of 70 laparoscopic Roux-en-Y gastric bypass videos, encompassing 303,764 frames. This dataset is fully incorporated into the pre-training stage to increase the procedural and anatomical diversity.
\item 
\textbf{StrasBypass70}~\cite{MultiBypass140} is a dataset dedicated to laparoscopic Roux-en-Y gastric bypass procedures, containing 70 surgical videos and a total of 457,787 frames. The entire dataset, comprising all available videos and frames, is used for pre-training to expand the diversity of surgical contexts.
\item 
\textbf{SurgWeb} is a web-sourced dataset constructed using the data collection pipeline proposed in the study~\cite{GSViT}. After manually removing noisy videos, the dataset consists of 3,376 videos with 2,349,618 frames, covering more than 20 distinct surgical procedures and related anatomical structures. The entire dataset is incorporated into the pre-training phase to enhance data diversity and strengthen the model’s spatiotemporal modeling capability.
\end{itemize}

\subsubsection*{Benchmark Datasets} 
In this study, we define datasets involving surgical procedures included in the pre-training data as in-domain, while those comprising previously unseen surgical procedures are referred to as out-of-domain. All evaluation datasets are sampled at 1 frame per second (1 FPS). Detailed information is provided in Extended Data Table~\ref{downstream}. 

The in-domain datasets include the following:
\begin{itemize}
\item 
\textbf{Cholec80}~\cite{endonet} is a benchmark dataset designed for surgical workflow recognition in laparoscopic cholecystectomy. Each frame is annotated with one of seven surgical workflow categories. 
Following the official data split protocol, our pre-training stage includes videos from both the Cholec80 and M2CAI16-Workflow training sets. However, we identify a possible data leakage issue due to five overlapping videos between M2CAI16-Workflow and the Cholec80 test set. To avoid any risk of data leakage and ensure fair evaluation, we conservatively exclude five overlapping videos from the Cholec80 test set, which are potentially included in the M2CAI16-Workflow training set. After excluding the five overlapping videos, we use 40 videos containing 86,344 frames for training, and the remaining 35 videos with 88,494 frames for testing. 
\item 
\textbf{Cholec80-CVS}~\cite{Cholec80-CVS} is a dataset for surgical skill assessment, providing video-level annotations of Strasberg’s Critical View of Safety (CVS) criteria for Cholec80. Each clip is labeled with three scores corresponding to the CVS criteria, with each score ranging from 0 to 2, resulting in a total score ranging from 0 to 6. The dataset is partitioned into training, validation and testing sets, while manually excluding samples labeled with a score of 6 that only appear in the test set. The training set consists of 263 clips totaling 17,201 frames, the validation set contains 43 clips totaling 3,327 frames, and the test set comprises 137 clips totaling 10,469 frames.

\item 
\textbf{CholecT50}~\cite{CholecT50} is a dataset designed for surgical triplet recognition in laparoscopic cholecystectomy. The dataset comprises 6 instruments, 10 verbs, and 15 targets, yielding a total of 100 triplet classes in the format of $\langle\text{instrument}, \text{verb}, \text{target}\rangle$. Following the common partition~\cite{CholecT50,CholecT50_split}, 45 videos containing 90,489 frames are designated for training, while the remaining 5 videos with 10,374 frames are reserved for testing.
\item 
\textbf{M2CAI16-Workflow}~\cite{endonet,m2cai16workflow} is a  benchmark dataset used for surgical workflow recognition 
in laparoscopic cholecystectomy. Each frame is labeled with one of eight distinct surgical workflow categories. Consistent with the standard partition, 27 videos containing 67,578 frames are allocated for training, and the remaining 14 videos comprising 26,961 frames are set aside for testing.
\item 
\textbf{Endoscapes-CVS}~\cite{Endoscapes-CVS201} is a dataset designed for surgical skill assessment in laparoscopic cholecystectomy. It comprises 201 clips, with each frame annotated using three averaged expert-assigned scores corresponding to CVS criteria. Each score is thresholded into a binary value (0 or 1), resulting in a total discrete score ranging from 0 to 3. A total of 120 videos, comprising 36,694 frames, are allocated for training, 41 videos containing 12,372 frames are used for validation, and the remaining 40 videos, with 9,747 frames, are designated for testing.
\item 
\textbf{PmLR50}~\cite{PmLR50} is a dataset for surgical workflow recognition in laparoscopic liver resection, containing 50 videos with totally 23,037 frames. Each frame is annotated with one of five distinct surgical workflow categories. Following the official partition, 35 videos totaling 17,378 frames are used for training, 5 videos with 2,309 frames are designated for validation, and 10 videos with 5,350 frames are reserved for testing.
\item 
\textbf{AutoLaparo}~\cite{AutoLaparo} is a dataset for image-guided surgical automation in laparoscopic hysterectomy. It contains 21 videos of hysterectomy procedures, comprising 83,243 frames. Each frame is annotated with one of seven distinct surgical workflow categories. Following the official partition, 10 videos totaling 40,211 frames are used for training, 4 videos with 14,972 frames are designated for validation, and 7 videos with 28,060 frames are reserved for testing.
\item 
\textbf{SurgicalActions160}~\cite{surgicalactions160} is a dataset containing 160 short-form clips, each depicting typical surgical actions in gynecologic laparoscopy. It consists of 16 distinct actions, with each action represented by exactly 10 examples. A total of 112 videos, comprising 13,238 frames, are allocated for training, while 48 videos with 5,783 frames are designated for testing.
\item 
\textbf{Prostate21} is a dataset consisting of 12 publicly available videos obtained from~\cite{PSI-AVA,ESAD} and 9 privately collected recordings, all capturing radical prostatectomy procedures. Based on manual annotations, the dataset includes 7 instruments, 10 verbs, and 10 targets, resulting in 89 unique triplet classes structured as $\langle\text{instrument}, \text{verb}, \text{target}\rangle$. To prevent potential data leakage, all videos originating from the PSI-AVA dataset~\cite{PSI-AVA} are allocated exclusively to the training set. In total, the training set consists of 14 videos with 38,730 frames, the validation set includes 3 videos with 9,571 frames, and the test set comprises 4 videos with 12,228 frames.
\end{itemize}

The out-of-domain datasets include the following:
\begin{itemize}
\item 
\textbf{CATARACTS}~\cite{Cataracts}  is a dataset designed for surgical workflow recognition in cataract surgery, consisting of 50 videos and 31,955 frames. In this dataset, the cataract surgery is divided into 19 phases and each frame is annotated with a phase label. Following the official partition, 25 videos totaling 16,507 frames are allocated for training, while 5 videos and 20 videos with 15,448 frames are designated for validation and testing.
\item 
\textbf{Cataract-101}~\cite{cataract-101} is a dataset designed for surgical workflow recognition in cataract surgery, comprising 101 untrimmed videos, of which 100 are employed for downstream task. Each frame is assigned to one of ten distinct surgical phases. The dataset is partitioned into 50 videos with 25,288 frames for training, 10 videos with 4,148 frames for validation, and 40 videos with 18,215 frames for testing.
\item 
\textbf{Cataract-21}~\cite{cataract21} is a dataset designed for surgical workflow recognition in cataract surgery. It consists of 21 video recordings, with the surgical procedure divided into 11 distinct phases, each frame annotated with the corresponding phase. The dataset includes 18 videos with 7,716 frames for training and 3 videos with 937 frames for testing.
\item 
\textbf{ESD57} is a private dataset designed for surgical workflow recognition in endoscopic submucosal dissection (ESD) procedures, containing 57 video recordings spanning the stomach and lower digestive tract. Each frame is labeled with one of eight distinct workflow phases: Lesion Detection, Lesion Marking, Submucosal Injection, Mucosal Incision, Submucosal Dissection, Wound Management, Wound Suturing, and Others. A total of 39 videos with 121,225 frames are used for training, 6 videos totaling 24,187 frames for validation, and the remaining 12 videos totaling 38,934 frames are allocated for testing.
\end{itemize}

\subsection*{Data availability}
The publicly available datasets used to construct the pre-training corpus and evaluation benchmark are summarized in Extended Data Table~\ref{data_available_training}.
\subsection*{Code availability}
The implementations of SurgVISTA framework will be released in GitHub: \href{https://github.com/isyangshu/SurgVISTA}{https://github.com/isyangshu/SurgVISTA}. The pre-trained natural-domain parameters used in this study are listed in \textbf{Extended Data Table~\ref{tab:natural_params}}, while the pre-trained surgical-domain parameters are listed in \textbf{Extended Data Table~\ref{tab:surgical_params}}. The other public codes used in this study are listed in \textbf{Extended Data Table~\ref{tab:public_code}}.
\subsection*{Ethics declarations}
This project has been reviewed and approved by the Human and Artefacts Research Ethics Committee (HAREC). The protocol number is HREP-2024-0379.
\subsection*{Author contribution}
S.Y., L.M.-H., and H.C. conceived and designed the work. S.Y. contributed to the technical implementation and conducted experiments.
F.Z. participated in discussions regarding the design of the self-supervised learning framework and were responsible for reproducing the natural-domain models. L.M. participated in discussions regarding the design of the self-supervised learning framework and contributed to part of the experimental evaluations. F.H., Y.W., S.H., Y.N., and Y.C. collected the data for self-supervised learning and downstream task evaluation. Xi.W., Ö.S., Y.J. and J.Q. offered insightful suggestions for the experimental design and thoughtfully directing the research trajectory. H.S., S.X., A.Q.L, Z.L., and J.Y.T. provided clinical expertise and facilitated access to proprietary datasets. All authors contributed to the drafting and revising of the manuscript. L.M.-H. and H.C. supervised the research.
\subsection*{Declarations}
The authors have no conflicts of interest to declare.
\subsection*{Acknowledgements}
The work described in this paper was supported by a grant from the Germany/Hong Kong Joint Research Scheme sponsored by the Research Grants Council of Hong Kong and the Germany Academic Exchange Service of Germany (Reference No. G-HKUST605/24) and Hong Kong Innovation and Technology Commission (Project No. GHP/006/22GD and ITCPD/17-9). 

\bibliography{sn-bibliography}
\newpage
\begin{appendices}
\section{Extended Data} \label{extended_data}

\begin{table*}[!h]
	\centering
	\caption{Detailed distribution of the SurgPub and SurgWeb datasets, detailing anatomical structures, surgical procedures, and their respective video and frame counts, offering a comprehensive overview of the dataset composition and organization.}
    \resizebox{0.9\textwidth}{!}{
	\begin{tabular}	{ c c  c c}
	\toprule
    \textbf{Anatomical Structure} & \textbf{Surgical Procedure} &  \textbf{Video Count} & \textbf{Frame Count} \\
    \midrule
    \rowcolor{dino}
    \multicolumn{4}{c}{\textbf{SurgPub}}\\
    \midrule
    Pituitary & Endoscopic Pituitary Surgery & 25 & 120018 \\
    \midrule
    Hepatobiliary &  Laparoscopic Cholecystectomy  & 91 & 209061 \\
    \midrule
    Stomach & Laparoscopic Roux-en-Y Gastric Bypass Surgery & 140 & 761551 \\
    \midrule
    Prostate & Radical Prostatectomy & 8 & 72318 \\
    \midrule
    Uterus & Laparoscopic Hysterectomy & 10 & 40211 \\
    \midrule
    \textbf{Total} & & \textbf{274} & \textbf{1203159} \\
    \midrule
    \rowcolor{dino}
    \multicolumn{4}{c}{\textbf{SurgWeb}}\\
    \midrule
    \multirow{2}{1.5cm}{\centering Hepatobiliary} & Laparoscopic Cholecystectomy & 375 & 326234 \\
    &Liver Resection & 75 & 67779 \\
    \midrule
    Spleen & Splenectomy & 30 & 31192 \\
    \midrule
    Abdominal Wall & Hernia Repair & 199 & 135038 \\
    \midrule
    Muscle & Myotomy & 25 & 76216 \\
    \midrule
    Appendix & Appendectomy & 98 & 59819 \\
    \midrule
    Kidney & Laparoscopic Nephrectomy & 8 & 5684 \\
    \midrule
    \multirow{2}{1.5cm}{\centering Esophagus} &  Esophagectomy & 20 & 50677 \\
    & Esophagojejunostomy & 59 & 5741 \\
    \midrule
    \multirow{4}{1.5cm}{\centering Stomach} &  Gastrectomy & 10 & 21940 \\
    & Gastrojejunostomy & 11 & 10124 \\
    & Fundoplication & 8 & 12730 \\
    & Others & 12 & 5253 \\
    \midrule
    \multirow{7}{1.5cm}{\centering {Lower  Digestive Tract}} &  Colectomy & 592 & 426677 \\
    & Rectopexy & 17 & 13340 \\
    & Sigmoidectomy & 14 & 14800 \\
    & Ladd's Procedure & 16 & 9212 \\
    & Rectal Cancer Surgery & 117 & 49956 \\
    & Anal Surgery & 50 & 17217 \\
    & Others & 86 & 35209 \\
    \midrule
    Others & Others & 1554 & 974770 \\
    \midrule
    \textbf{Total} & & \textbf{3376} & \textbf{2349618} \\
    \midrule
    \rowcolor{dino}
    \multicolumn{4}{c}{\textbf{Total}}\\
    \midrule
    \textbf{Total} & & \textbf{3650} & \textbf{3552777} \\
    \bottomrule
	\end{tabular}}
	\label{lab:SurgPubandSurgWeb}
\end{table*}
\begin{table}[htbp]
\centering
\begin{turn}{90}
\begin{minipage}{\textheight}
\centering
\caption{Detailed composition of the large-scale video-level pre-training dataset, consisting of eight publicly available surgical video datasets spanning five distinct procedures, alongside one dataset containing surgical videos sourced from online platforms. The ``Statistic'' column presents the counts of videos (in \textcolor{blue}{Blue}) and frames (in \textcolor{red}{Red}) for both training and test splits. The ``Available'' column denotes the subset of data (videos, frames, and average duration) actually used during pre-training to avoid data leakage (e.g., inclusion of test sets or duplicate samples). The ``Pre-training Data Scale Variations'' column outlines the different data configurations adopted in various pre-training settings.}
\label{lab:pre-training_dataset}
\resizebox{\linewidth}{!}{
\begin{tabular}{ c c c c c c c c c c c c c c }
\toprule
     \multirow{2}{0.5cm}{\centering \textbf{\#}} & \multirow{2}{5.5cm}{\centering \textbf{Dataset}} & \multirow{2}{3cm}{\centering \textbf{Surgical Procedure}} &  \multicolumn{3}{c}{\textbf{Statistic}} & \multicolumn{3}{c}{\textbf{Available}} & \multicolumn{5}{c}{\textbf{Pre-training Data Scale Variations}} \\
     \cmidrule(lr){4-6} \cmidrule(lr){7-9} \cmidrule(lr){10-14}
     &&& Train & Test & All & Video & Frame & Average & A (86.3k) & B (153.9k) & C (441.6k) & D (1.20M) & E (3.55M) \\
    \midrule
    \multirow{2}{0.5cm}{\centering \textbf{1}} & \multirow{2}{5.5cm}{\centering Cholec80~\cite{endonet}} & \multirow{2}{3cm}{\centering Laparoscopic \\ Cholecystectomy}  & \multirow{2}{1.2cm}{\centering \textcolor{blue}{40} \\ (\textcolor{red}{86344})} & \multirow{2}{1.2cm}{\centering \textcolor{blue}{40} \\ (\textcolor{red}{98234})} & \multirow{2}{1.2cm}{\centering \textcolor{blue}{80} \\ (\textcolor{red}{184578})}  & \multirow{2}{0.5cm}{\centering 40} & \multirow{2}{0.5cm}{\centering 86344} & \multirow{2}{0.5cm}{\centering 2158} & \multirow{2}{0.2cm}{\Checkmark} & \multirow{2}{0.2cm}{\Checkmark} & \multirow{2}{0.2cm}{\Checkmark} & \multirow{2}{0.2cm}{\Checkmark} & \multirow{2}{0.2cm}{\Checkmark} \\
    &&&&&&&\\
    \midrule
    \multirow{2}{0.5cm}{\centering \textbf{2}} & \multirow{2}{5.5cm}{\centering MI2CAI16-Workflow~\cite{endonet,m2cai16workflow}} & \multirow{2}{3cm}{\centering Laparoscopic \\ Cholecystectomy}  & \multirow{2}{1.2cm}{\centering \textcolor{blue}{27} \\ (\textcolor{red}{67578})} & \multirow{2}{1.2cm}{\centering \textcolor{blue}{14} \\ (\textcolor{red}{26961})} & \multirow{2}{1.2cm}{\centering \textcolor{blue}{41} \\ (\textcolor{red}{94539})}  & \multirow{2}{0.5cm}{\centering 27} & \multirow{2}{0.5cm}{\centering 67578} & \multirow{2}{0.5cm}{\centering 2502} & & \multirow{2}{0.2cm}{\Checkmark} & \multirow{2}{0.2cm}{\Checkmark} & \multirow{2}{0.2cm}{\Checkmark} & \multirow{2}{0.2cm}{\Checkmark} \\
    &&&&&&&\\
    \midrule
    \multirow{2}{0.5cm}{\centering \textbf{3}} & \multirow{2}{5.5cm}{\centering HeiChole~\cite{HeiChole}} & \multirow{2}{3cm}{\centering Laparoscopic \\ Cholecystectomy}  & \multirow{2}{1.2cm}{\centering \textcolor{blue}{24} \\ (\textcolor{red}{55139})} & \multirow{2}{1.2cm}{\centering \textcolor{blue}{9} \\ (-)} & \multirow{2}{1.2cm}{\centering \textcolor{blue}{33} \\ (-)}  & \multirow{2}{0.5cm}{\centering 24} & \multirow{2}{0.5cm}{\centering 55139} & \multirow{2}{0.5cm}{\centering 2297} &&& \multirow{2}{0.2cm}{\Checkmark} & \multirow{2}{0.2cm}{\Checkmark} & \multirow{2}{0.2cm}{\Checkmark}\\
    &&&&&&&\\
    \midrule
    \multirow{2}{0.5cm}{\centering \textbf{4}} & \multirow{2}{5.5cm}{\centering PitVis~\cite{PitVis}} & \multirow{2}{3.0cm}{\centering Endoscopic \\ Pituitary Surgery}  & \multirow{2}{1.2cm}{\centering \textcolor{blue}{25} \\ (\textcolor{red}{120018})} & \multirow{2}{1.2cm}{\centering \textcolor{blue}{8} \\ (-)} & \multirow{2}{1.2cm}{\centering \textcolor{blue}{33} \\ (-)}  & \multirow{2}{0.5cm}{\centering 25} & \multirow{2}{0.5cm}{\centering 120018} & \multirow{2}{0.5cm}{\centering 4800} && & \multirow{2}{0.2cm}{\Checkmark} & \multirow{2}{0.2cm}{\Checkmark} & \multirow{2}{0.2cm}{\Checkmark} \\
    &&&&&&&\\
    \midrule
    \multirow{2}{0.5cm}{\centering \textbf{5}} & \multirow{2}{5.5cm}{\centering PSI-AVA~\cite{PSI-AVA}} & \multirow{2}{3.0cm}{\centering Radical \\ Prostatectomy}  & \multirow{2}{1.2cm}{\centering - \\ (-)} & \multirow{2}{1.2cm}{\centering - \\ (-)} & \multirow{2}{1.2cm}{\centering \textcolor{blue}{8} \\ (\textcolor{red}{73626})}  & \multirow{2}{0.5cm}{\centering 8} & \multirow{2}{0.5cm}{\centering 72318} & \multirow{2}{0.5cm}{\centering 9039} && & \multirow{2}{0.2cm}{\Checkmark} & \multirow{2}{0.2cm}{\Checkmark} & \multirow{2}{0.2cm}{\Checkmark}\\
    &&&&&&&\\
    \midrule
    \multirow{2}{0.5cm}{\centering \textbf{6}} & \multirow{2}{5.5cm}{\centering AutoLaparo~\cite{AutoLaparo}} & \multirow{2}{3.0cm}{\centering Laparoscopic \\ Hysterectomy}  & \multirow{2}{1.2cm}{\centering \textcolor{blue}{10} \\ (\textcolor{red}{40211})} & \multirow{2}{1.2cm}{\centering \textcolor{blue}{11} \\ (\textcolor{red}{43032})} & \multirow{2}{1.2cm}{\centering \textcolor{blue}{21} \\ (\textcolor{red}{83243})}  & \multirow{2}{0.5cm}{\centering 10} & \multirow{2}{0.5cm}{\centering 40211} & \multirow{2}{0.5cm}{\centering 4021} && & \multirow{2}{0.2cm}{\Checkmark} & \multirow{2}{0.2cm}{\Checkmark} & \multirow{2}{0.2cm}{\Checkmark} \\
    &&&&&&&\\
    \midrule
    \multirow{3}{0.5cm}{\centering \textbf{7}} & \multirow{3}{5.5cm}{\centering BernBypass70~\cite{MultiBypass140}} & \multirow{3}{3.0cm}{\centering Laparoscopic \\ Roux-en-Y Gastric Bypass Surgery}  & \multirow{3}{1.2cm}{\centering - \\ (-)} & \multirow{3}{1.2cm}{\centering - \\ (-)} & \multirow{3}{1.2cm}{\centering \textcolor{blue}{70} \\ (\textcolor{red}{316595})}  & \multirow{3}{0.5cm}{\centering 70} & \multirow{3}{0.5cm}{\centering 303764} & \multirow{3}{0.5cm}{\centering 4339} &&&& \multirow{3}{0.2cm}{\Checkmark} & \multirow{2}{0.2cm}{\Checkmark} \\
    &&&&&&&\\
    &&&&&&&\\
    \midrule
    \multirow{3}{0.5cm}{\centering \textbf{8}} & \multirow{3}{5.5cm}{\centering StrasBypass70~\cite{MultiBypass140}} & \multirow{3}{3.0cm}{\centering Laparoscopic \\ Roux-en-Y Gastric Bypass Surgery}  & \multirow{3}{1.2cm}{\centering - \\ (-)} & \multirow{3}{1.2cm}{\centering - \\ (-)} & \multirow{3}{1.2cm}{\centering \textcolor{blue}{70} \\ (\textcolor{red}{464989})}  & \multirow{3}{0.5cm}{\centering 70} & \multirow{3}{0.5cm}{\centering 457787} & \multirow{3}{0.5cm}{\centering 6540} &&&&\multirow{3}{0.2cm}{\Checkmark} & \multirow{2}{0.2cm}{\Checkmark} \\
    &&&&&&&\\
    &&&&&&&\\
    \midrule
    \multirow{2}{0.5cm}{\centering \textbf{9}} & \multirow{2}{5.5cm}{\centering SurgWeb} & \multirow{2}{3.0cm}{\centering Hybrid Surgical Procedures}  & \multirow{2}{1.2cm}{\centering - \\ (-)} & \multirow{2}{1.2cm}{\centering - \\ (-)} & \multirow{2}{1.2cm}{\centering \textcolor{blue}{3376} \\ (\textcolor{red}{2349618})}  & \multirow{2}{0.5cm}{\centering 3376} & \multirow{2}{0.5cm}{\centering 2349618} & \multirow{2}{0.5cm}{\centering 696} &&&& & \multirow{2}{0.2cm}{\Checkmark}\\
    &&&&&&&\\
    \bottomrule
\end{tabular}}
\end{minipage}
\end{turn}
\end{table}
\begin{table*}[h]
	\centering
	\caption{Detailed composition of the comprehensive video-level downstream evaluation, encompassing 13 datasets spanning six surgical procedures across four distinct video-level tasks. Datasets are categorized as \textbf{in-domain} or \textbf{out-of-domain} depending on whether the corresponding surgical procedures are included in the pre-training data, thereby facilitating the assessment of both within-domain generalization and cross-procedural transferability.}

    \resizebox{0.95\textwidth}{!}{
\begin{tabular}	{ c c  c c c  c }
	\toprule
    \multirow{2}{4cm}{\centering \textbf{Dataset}} & \multirow{2}{3cm}{\centering \textbf{Surgical Procedure}} &  \multicolumn{3}{c}{\textbf{Statistic}} & \multirow{2}{3cm}{\centering \textbf{Task}} \\
     \cmidrule(lr){3-5} 
     && Train & Val+Test & All & \\
    \midrule
    \rowcolor{dino}
    \multicolumn{6}{c}{\textbf{In-domain Datasets}}\\
    \midrule
    \multirow{2}{4cm}{\centering Cholec80~\cite{endonet}} & \multirow{2}{3cm}{\centering Laparoscopic \\ Cholecystectomy}  & \multirow{2}{1.5cm}{\centering 40 \\ (86344)} & \multirow{2}{1.5cm}{\centering 35 \\ (88494)} & \multirow{2}{1.5cm}{\centering 75 \\ (174838)}  & \multirow{2}{3cm}{\centering Surgical Workflow Recognition} \\
    &&&&\\
    \midrule
    \multirow{2}{4cm}{\centering Cholec80-CVS~\cite{Cholec80-CVS}} & \multirow{2}{3cm}{\centering Laparoscopic \\ Cholecystectomy}  & \multirow{2}{1.5cm}{\centering 263 \\ (17201)} & \multirow{2}{1.5cm}{\centering 180 \\ (13796)} & \multirow{2}{1.5cm}{\centering 443 \\ (30997)}  & \multirow{2}{4cm}{\centering Surgical Skill Assessment} \\
    &&&&\\
    \midrule
    \multirow{2}{4cm}{\centering CholecT50~\cite{CholecT50}} & \multirow{2}{3cm}{\centering Laparoscopic \\ Cholecystectomy}  & \multirow{2}{1.5cm}{\centering 45 \\ (90489)} & \multirow{2}{1.5cm}{\centering 5 \\ (10374)} & \multirow{2}{1.5cm}{\centering 50 \\ (100863)} & \multirow{2}{3cm}{\centering Surgical Triplet Recognition} \\
\bottomrule
	\end{tabular}}
	\label{downstream}
\end{table*}

\begin{table*}[t]
    \centering
    \caption{Overall performance evaluation of SurgVISTA against state-of-the-art image- and video-level pre-training methods within the natural domain on the Cholec80 dataset.}
    \resizebox{0.78\textwidth}{!}{
    \begin{tabular}	{ c c c c c c c c}
	\toprule
    \multirow{2}{*}{\textbf{Method}} & \multirow{2}{*}{\textbf{Pretraining Data}} & \textbf{Image-level} & \textbf{Video-level} & \multicolumn{3}{c}{\textbf{Phase-level}} \\
    \cmidrule(lr){3-3} \cmidrule(lr){4-4} \cmidrule(lr){5-7}
    & & \textbf{Accuracy $\uparrow$} & \textbf{Accuracy $\uparrow$} & \textbf{Precision $\uparrow$} & \textbf{Recall $\uparrow$} & \textbf{Jaccard $\uparrow$} \\
    \midrule
    \rowcolor{dino}
    \multicolumn{7}{c}{\textbf{Image-level}}\\
    \midrule
    \multirow{2}{*}{Supervised~\cite{ViT}} & \multirow{2}{*}{ImageNet~\cite{imagenet}} &
    87.5 & 87.9 & 84.7 & 83.6 & 71.9 \\
    &&(87.2, 87.7) & (87.7, 88.2) & (84.3, 85.1) & (83.2, 84.0) & (71.4, 72.4) \\
    \midrule
    \multirow{2}{*}{DINO~\cite{DINO}} & \multirow{2}{*}{ImageNet~\cite{imagenet}} & 
   86.9 & 87.2 & 83.4 & 81.5 & 69.6 \\
   && (86.7, 87.1) & (87.0, 87.4) & (82.9, 83.8) & (81.0, 81.9) & (69.0, 70.2) \\
   \midrule
    \multirow{2}{*}{MAE~\cite{MAE}} &  \multirow{2}{*}{ImageNet~\cite{imagenet}} & 
    88.0 & 88.3 & 84.4 & 83.0 & 71.4 \\
\bottomrule
	\end{tabular}}
    \label{lab:Cholec80}
\end{table*}

\begin{table*}[t]
    \centering
    \caption{    Overall performance evaluation of SurgVISTA against state-of-the-art image- and video-level pre-training methods within the natural domain on the M2CAI16-Workflow dataset.}
    \resizebox{0.78\textwidth}{!}{
    \begin{tabular}	{ c c c c c c c c}
	\toprule
    \multirow{2}{*}{\textbf{Method}} & \multirow{2}{*}{\textbf{Pretraining Data}} & \textbf{Image-level} & \textbf{Video-level} & \multicolumn{3}{c}{\textbf{Phase-level}} \\
    \cmidrule(lr){3-3} \cmidrule(lr){4-4} \cmidrule(lr){5-7}
    & & \textbf{Accuracy $\uparrow$} & \textbf{Accuracy $\uparrow$} & \textbf{Precision $\uparrow$} & \textbf{Recall $\uparrow$} & \textbf{Jaccard $\uparrow$} \\
    \midrule
    \rowcolor{dino}
    \multicolumn{7}{c}{\textbf{Image-level}}\\
    \midrule
    \multirow{2}{*}{Supervised~\cite{ViT}} & \multirow{2}{*}{ImageNet~\cite{imagenet}} &
    80.0 & 81.0 & 79.3 & 78.3 & 63.0 \\
&&(79.5, 80.5) & (80.5, 81.5) & (78.7, 79.9) & (77.6, 79.0) & (62.1, 63.8) \\
\midrule
    \multirow{2}{*}{DINO~\cite{DINO}} & \multirow{2}{*}{ImageNet~\cite{imagenet}} & 
   77.2 & 78.8 & 77.2 & 76.1 & 60.7 \\
&&(76.7, 77.7) & (78.3, 79.3) & (76.5, 77.9) & (75.4, 76.8) & (59.9, 61.5) \\
\midrule
    \multirow{2}{*}{MAE~\cite{MAE}} &  \multirow{2}{*}{ImageNet~\cite{imagenet}} & 
    78.9 & 81.1 & 79.5 & 78.9 & 63.8 \\
\bottomrule
	\end{tabular}}
    \label{lab:M2CAI16-Workflow}
\end{table*}

\begin{table*}[t]
    \centering
    \caption{Overall performance evaluation of SurgVISTA against state-of-the-art image- and video-level pre-training methods within the natural domain on the PmLR50 dataset.}
    \resizebox{0.78\textwidth}{!}{
    \begin{tabular}	{ c c c c c c c c}
	\toprule
    \multirow{2}{*}{\textbf{Method}} & \multirow{2}{*}{\textbf{Pretrained Data}} & \textbf{Image-level} & \textbf{Video-level} & \multicolumn{3}{c}{\textbf{Phase-level}} \\
    \cmidrule(lr){3-3} \cmidrule(lr){4-4} \cmidrule(lr){5-7}
    & & \textbf{Accuracy $\uparrow$} & \textbf{Accuracy $\uparrow$} & \textbf{Precision $\uparrow$} & \textbf{Recall $\uparrow$} & \textbf{Jaccard $\uparrow$} \\
    \midrule
    \rowcolor{dino}
    \multicolumn{7}{c}{\textbf{Image-level}}\\
    \midrule
    \multirow{2}{*}{Supervised~\cite{ViT}} & \multirow{2}{*}{ImageNet~\cite{imagenet}} &
    79.6 & 78.9 & 81.0 & 59.6 & 48.8 \\
&&(78.6, 80.6) & (77.9, 80.0) & (78.8, 83.2) & (57.8, 61.3) & (46.7, 50.8) \\
\midrule
    \multirow{2}{*}{DINO~\cite{DINO}} & \multirow{2}{*}{ImageNet~\cite{imagenet}} & 
  86.5 & 86.1 & 88.5 & 72.7 & 64.5 \\
&&(85.6, 87.4) & (85.1, 87.0) & (87.2, 89.8) & (70.7, 74.7) & (62.2, 66.8) \\
\midrule
    \multirow{2}{*}{MAE~\cite{MAE}} &  \multirow{2}{*}{ImageNet~\cite{imagenet}} & 
   86.9 & 86.3 & 83.9 & 72.0 & 62.1 \\
\bottomrule
	\end{tabular}}
    \label{lab:PmLR50}
\end{table*}

\begin{table*}[t]
    \centering
    \caption{Overall performance evaluation of SurgVISTA against state-of-the-art image- and video-level pre-training methods within the natural domain on the AutoLaparo dataset.}
    \resizebox{0.78\textwidth}{!}{
    \begin{tabular}	{ c c c c c c c c}
	\toprule
    \multirow{2}{*}{\textbf{Method}} & \multirow{2}{*}{\textbf{Pretraining Data}} & \textbf{Image-level} & \textbf{Video-level} & \multicolumn{3}{c}{\textbf{Phase-level}} \\
    \cmidrule(lr){3-3} \cmidrule(lr){4-4} \cmidrule(lr){5-7}
    & & \textbf{Accuracy $\uparrow$} & \textbf{Accuracy $\uparrow$} & \textbf{Precision $\uparrow$} & \textbf{Recall $\uparrow$} & \textbf{Jaccard $\uparrow$} \\
    \midrule
    \rowcolor{dino}
    \multicolumn{7}{c}{\textbf{Image-level}}\\
    \midrule
    \multirow{2}{*}{Supervised~\cite{ViT}} & \multirow{2}{*}{ImageNet~\cite{imagenet}} &
    80.1 & 80.5 & 76.0 & 68.8 & 58.6 \\
&&(79.7, 80.6) & (80.0, 80.9) & (67.3, 84.7) & (67.6, 70.1) & (57.4, 59.8) \\
\midrule
    \multirow{2}{*}{DINO~\cite{DINO}} & \multirow{2}{*}{ImageNet~\cite{imagenet}} & 
   81.7 & 82.0 & 81.2 & 78.6 & 66.1 \\
&&(81.2, 82.1) & (81.5, 82.4) & (79.4, 83.0) & (77.1, 80.2) & (64.6, 67.6) \\
\midrule
    \multirow{2}{*}{MAE~\cite{MAE}} &  \multirow{2}{*}{ImageNet~\cite{imagenet}} & 
    77.0 & 77.2 & 74.4 & 68.3 & 56.6 \\
\bottomrule
	\end{tabular}}
    \label{lab:AutoLaparo}
\end{table*}

\begin{table*}[t]
    \centering
    \caption{Overall performance evaluation of SurgVISTA against state-of-the-art image- and video-level pre-training methods within the natural domain on the ESD57 dataset.}

    \resizebox{0.78\textwidth}{!}{
    \begin{tabular}	{ c c c c c c c c}
	\toprule
    \multirow{2}{*}{\textbf{Method}} & \multirow{2}{*}{\textbf{Pretrained Data}} & \textbf{Image-level} & \textbf{Video-level} & \multicolumn{3}{c}{\textbf{Phase-level}} \\
    \cmidrule(lr){3-3} \cmidrule(lr){4-4} \cmidrule(lr){5-7}
    & & \textbf{Accuracy $\uparrow$} & \textbf{Accuracy $\uparrow$} & \textbf{Precision $\uparrow$} & \textbf{Recall $\uparrow$} & \textbf{Jaccard $\uparrow$} \\
    \midrule
    \rowcolor{dino}
    \multicolumn{7}{c}{\textbf{Image-level}}\\
    \midrule
    \multirow{2}{*}{Supervised~\cite{ViT}} & \multirow{2}{*}{ImageNet~\cite{imagenet}} &
    66.1 & 69.1 & 75.7 & 62.0 & 50.2 \\
&&(65.7, 66.6) & (68.6, 69.6) & (74.7, 76.8) & (61.0, 63.0) & (49.3, 51.1) \\
\midrule
    \multirow{2}{*}{DINO~\cite{DINO}} & \multirow{2}{*}{ImageNet~\cite{imagenet}} & 
  68.2 & 70.8 & 76.1 & 63.9 & 51.6 \\
&&(67.8, 68.7) & (70.4, 71.3) & (75.1, 77.1) & (62.7, 65.0) & (50.6, 52.6) \\
\midrule
    \multirow{2}{*}{MAE~\cite{MAE}} &  \multirow{2}{*}{ImageNet~\cite{imagenet}} & 
   67.4 & 70.1 & \textbf{77.8} & 62.0 & 49.7 \\
\bottomrule
	\end{tabular}}
    \label{lab:ESD57}
\end{table*}

\begin{table*}[t]
    \centering
    \caption{Overall performance evaluation of SurgVISTA against state-of-the-art image- and video-level pre-training methods within the natural domain on the CATARACTS dataset.}
    \resizebox{0.78\textwidth}{!}{
    \begin{tabular}	{ c c c c c c c c}
	\toprule
    \multirow{2}{*}{\textbf{Method}} & \multirow{2}{*}{\textbf{Pretraining Data}} & \textbf{Image-level} & \textbf{Video-level} & \multicolumn{3}{c}{\textbf{Phase-level}} \\
    \cmidrule(lr){3-3} \cmidrule(lr){4-4} \cmidrule(lr){5-7}
    & & \textbf{Accuracy $\uparrow$} & \textbf{Accuracy $\uparrow$} & \textbf{Precision $\uparrow$} & \textbf{Recall $\uparrow$} & \textbf{Jaccard $\uparrow$} \\
    \midrule
    \rowcolor{dino}
    \multicolumn{7}{c}{\textbf{Image-level}}\\
    \midrule
    \multirow{2}{*}{Supervised~\cite{ViT}} & \multirow{2}{*}{ImageNet~\cite{imagenet}} &
    74.5 & 78.0 & 67.0 & 63.0 & 52.5 \\
&&(73.8, 75.3) & (77.3, 78.8) & (65.9, 68.2) & (61.9, 64.2) & (51.2, 53.8) \\
\midrule
    \multirow{2}{*}{DINO~\cite{DINO}} & \multirow{2}{*}{ImageNet~\cite{imagenet}} & 
   80.7 & 84.9 & 69.1 & 67.2 & 59.6 \\
&&(80.0, 81.3) & (84.3, 85.5) & (68.6, 69.6) & (66.3, 68.1) & (58.6, 60.7) \\
\midrule
    \multirow{2}{*}{MAE~\cite{MAE}} &  \multirow{2}{*}{ImageNet~\cite{imagenet}} & 
    81.1 & 84.5 & 74.0 & 68.7 & 61.6 \\
\bottomrule
	\end{tabular}}
    \label{lab:CATARACT}
\end{table*}

\begin{table*}[t]
    \centering
    \caption{Overall performance evaluation of SurgVISTA against state-of-the-art image- and video-level pre-training methods within the natural domain on the Cataract-101 dataset.}

    \resizebox{0.78\textwidth}{!}{
    \begin{tabular}	{ c c c c c c c c}
	\toprule
    \multirow{2}{*}{\textbf{Method}} & \multirow{2}{*}{\textbf{Pretraining Data}} & \textbf{Image-level} & \textbf{Video-level} & \multicolumn{3}{c}{\textbf{Phase-level}} \\
    \cmidrule(lr){3-3} \cmidrule(lr){4-4} \cmidrule(lr){5-7}
    & & \textbf{Accuracy $\uparrow$} & \textbf{Accuracy $\uparrow$} & \textbf{Precision $\uparrow$} & \textbf{Recall $\uparrow$} & \textbf{Jaccard $\uparrow$} \\
    \midrule
    \rowcolor{dino}
    \multicolumn{7}{c}{\textbf{Image-level}}\\
    \midrule
    \multirow{2}{*}{Supervised~\cite{ViT}} & \multirow{2}{*}{ImageNet~\cite{imagenet}} &
    89.9 & 90.6 & 88.1 & 88.7 & 79.2 \\
&&(89.5, 90.3) & (90.2, 91.1) & (87.4, 88.7) & (88.1, 89.3) & (78.4, 80.1) \\
\midrule
    \multirow{2}{*}{DINO~\cite{DINO}} & \multirow{2}{*}{ImageNet~\cite{imagenet}} & 
  74.5 & 75.3 & 72.6 & 68.8 & 53.8 \\
&&(73.9, 75.1) & (74.6, 76.0) & (71.5, 73.7) & (67.9, 69.7) & (52.8, 64.7) \\
\midrule
    \multirow{2}{*}{MAE~\cite{MAE}} &  \multirow{2}{*}{ImageNet~\cite{imagenet}} & 
   81.8 & 82.4 & 77.9 & 77.4 & 63.8 \\
\bottomrule
	\end{tabular}}
    \label{lab:Cataract-101}
\end{table*}

\begin{table*}[t]
    \centering
    \caption{Overall performance evaluation of SurgVISTA against state-of-the-art image- and video-level pre-training methods within the natural domain on the Cataract-21 dataset.}
    \resizebox{0.78\textwidth}{!}{
    \begin{tabular}	{ c c c c c c c c}
	\toprule
    \multirow{2}{*}{\textbf{Method}} & \multirow{2}{*}{\textbf{Pretraining Data}} & \textbf{Image-level} & \textbf{Video-level} & \multicolumn{3}{c}{\textbf{Phase-level}} \\
    \cmidrule(lr){3-3} \cmidrule(lr){4-4} \cmidrule(lr){5-7}
    & & \textbf{Accuracy $\uparrow$} & \textbf{Accuracy $\uparrow$} & \textbf{Precision $\uparrow$} & \textbf{Recall $\uparrow$} & \textbf{Jaccard $\uparrow$} \\
    \midrule
    \rowcolor{dino}
    \multicolumn{7}{c}{\textbf{Image-level}}\\
    \midrule
    \multirow{2}{*}{Supervised~\cite{ViT}} & \multirow{2}{*}{ImageNet~\cite{imagenet}} &
    84.7 & 83.6 & 77.5 & 77.9 & 63.6 \\
&&(82.4, 86.9) & (81.2, 86.0) & (74.0, 80.9) & (74.8, 80.9) & (59.4, 67.8) \\
\midrule
    \multirow{2}{*}{DINO~\cite{DINO}} & \multirow{2}{*}{ImageNet~\cite{imagenet}} & 
  90.0 & 89.6 & 84.0 & 84.7 & 72.9 \\
&&(88.1, 91.9) & (87.6, 91.6) & (80.5, 87.5) & (81.7, 87.6) & (68.5, 77.3) \\
\midrule
    \multirow{2}{*}{MAE~\cite{MAE}} &  \multirow{2}{*}{ImageNet~\cite{imagenet}} & 
   87.2 & 86.4 & 82.0 & 81.1 & 69.6 \\
\bottomrule
	\end{tabular}}
    \label{lab:Cataract-21}
\end{table*}

\begin{table*}[t]
    \centering
    \caption{Overall performance evaluation of SurgVISTA against state-of-the-art image- and video-level pre-training methods within the natural domain on the CholecT50 dataset.}
    \resizebox{0.85\textwidth}{!}{
    \begin{tabular}	{ c c c c c c c c c}
	\toprule
    \multirow{2}{*}{\textbf{Method}} & \multirow{2}{*}{\textbf{Pretraining Data}} & \multicolumn{3}{c}{\textbf{Component detection}} & \multicolumn{3}{c}{\textbf{Triplet association}} \\
    \cmidrule(lr){3-5} \cmidrule(lr){6-8}
    & & \textbf{$AP_I$} & \textbf{$AP_v$} & \textbf{$AP_t$} & \textbf{$AP_{iv}$} & \textbf{$AP_{it}$} & \textbf{$AP_{ivt}$} \\
    \midrule
    \rowcolor{dino}
    \multicolumn{8}{c}{\textbf{Image-level}}\\
    \midrule
    \multirow{2}{*}{Supervised~\cite{ViT}} & \multirow{2}{*}{ImageNet~\cite{imagenet}} &
    80.7 & 49.1 & 41.4 & 33.7 & 34.5 & 29.8 \\
&&(78.9, 82.5) & (46.7, 51.5) & (37.8, 45.0) & (30.9, 36.5) & (31.9, 37.0) & (27.3, 32.3) \\
\midrule
    \multirow{2}{*}{DINO~\cite{DINO}} & \multirow{2}{*}{ImageNet~\cite{imagenet}} & 
    81.8 & 50.9 & 41.3 & 35.1 & 38.7 & 35.2 \\
&&(80.3, 83.2) & (48.6, 53.1) & (37.9, 44.7) & (32.1, 38.0) & (36.0, 41.3) & (32.5, 38.0) \\
\midrule
    \multirow{2}{*}{MAE~\cite{MAE}} &  \multirow{2}{*}{ImageNet~\cite{imagenet}} & 
    \textbf{85.6} & 52.4 & 43.0 & 37.7 & 38.8 & 35.3 \\
\bottomrule
	\end{tabular}}
    \label{lab:CholecT50}
\end{table*}

\begin{table*}[t]
    \centering
    \caption{Overall performance evaluation of SurgVISTA against state-of-the-art image- and video-level pre-training methods within the natural domain on the Prostate21 dataset.}
    \resizebox{0.85\textwidth}{!}{
    \begin{tabular}	{ c c c c c c c c c}
	\toprule
    \multirow{2}{*}{\textbf{Method}} & \multirow{2}{*}{\textbf{Pretraining Data}} & \multicolumn{3}{c}{\textbf{Component detection}} & \multicolumn{3}{c}{\textbf{Triplet association}} \\
    \cmidrule(lr){3-5} \cmidrule(lr){6-8}
    & & \textbf{$AP_I$} & \textbf{$AP_v$} & \textbf{$AP_t$} & \textbf{$AP_{iv}$} & \textbf{$AP_{it}$} & \textbf{$AP_{ivt}$} \\
    \midrule
    \rowcolor{dino}
    \multicolumn{8}{c}{\textbf{Image-level}}\\
    \midrule
    \multirow{2}{*}{Supervised~\cite{ViT}} & \multirow{2}{*}{ImageNet~\cite{imagenet}} &
    49.0 & 33.4 & 37.2 & 20.3 & 18.2 & 12.7 \\
&&(47.1, 50.8) & (32.4, 34.4) & (35.6, 38.9) & (19.5, 21.1) & (17.5, 18.9) & (12.1, 13.2) \\
\midrule
    \multirow{2}{*}{DINO~\cite{DINO}} & \multirow{2}{*}{ImageNet~\cite{imagenet}} & 
    69.3 & 55.0 & 48.8 & 36.1 & 31.3 & 21.3 \\
&&(67.1, 71.5) & (53.6, 56.5) & (46.7, 50.9) & (34.9, 37.3) & (30.0, 32.7) & (20.3, 22.2) \\
\midrule
    \multirow{2}{*}{MAE~\cite{MAE}} &  \multirow{2}{*}{ImageNet~\cite{imagenet}} & 
    62.9 & 50.2 & 47.8 & 33.0 & 29.6 & 21.2 \\
\bottomrule
	\end{tabular}}
    \label{lab:Prostate21}
\end{table*}
\begin{table*}[t]
    \centering
    \caption{Overall performance evaluation of SurgVISTA against state-of-the-art image- and video-level pre-training methods within the natural domain on the SurgicalActions160, Cholec80-CVS and Endoscapes-CVS datasets.}
    \resizebox{0.85\textwidth}{!}{
    \begin{tabular}	{ c c c c c c c c}
	\toprule
    \multirow{2}{*}{\textbf{Method}} & \multirow{2}{*}{\textbf{Pretraining Data}} & \multicolumn{2}{c}{\textbf{SurgicalActions160}} & \multicolumn{2}{c}{\textbf{Cholec80-CVS}} & \multicolumn{2}{c}{\textbf{Endoscapes-CVS}}\\
    \cmidrule(lr){3-4} \cmidrule(lr){5-6} \cmidrule(lr){7-8}
    & & \textbf{Acc} & \textbf{mAP} & \textbf{Acc} & \textbf{mAP} & \textbf{Acc} & \textbf{mAP} \\
    \midrule
    \rowcolor{dino}
    \multicolumn{8}{c}{\textbf{Image-level}}\\
    \midrule
    \multirow{2}{*}{Supervised~\cite{ViT}} & \multirow{2}{*}{ImageNet~\cite{imagenet}} &
    54.3 & 63.2 & 53.2 & 29.0 & 45.4 & 39.7 \\
    &&(40.9, 67.6) & (49.9, 76.5) & (44.9, 61.5) & (25.3, 32.8) & (44.5, 46.2) & (38.8, 40.6) \\
    \midrule
    \multirow{2}{*}{DINO~\cite{DINO}} & \multirow{2}{*}{ImageNet~\cite{imagenet}} & 
    60.4 & 60.2 & 44.4 & 20.8 & 44.7 & 40.9 \\
&&(46.7, 74.2) & (46.9, 73.4) & (35.9, 52.8) & (17.0, 24.5) & (43.8, 45.5) & (40.0, 41.7) \\
\midrule
    \multirow{2}{*}{MAE~\cite{MAE}} &  \multirow{2}{*}{ImageNet~\cite{imagenet}} & 
   64.8 & 63.1 & 45.9 & 26.0 & 45.7 & 41.3 \\
\bottomrule
	\end{tabular}}
    \label{lab:SurgicalActions160}
    \label{lab:Cholec80-CSV}
\end{table*}
\begin{table*}[t]
    \centering
    \caption{Overall performance comparison between SurgVISTA and state-of-the-art surgical pre-training methods on the Cholec80 dataset.}
    \resizebox{0.92\textwidth}{!}{
    \begin{tabular}	{ c c c c c c c c}
	\toprule
    \multirow{2}{*}{\textbf{Method}} & \multirow{2}{*}{\textbf{Pretraining Data}} & \textbf{Image-level} & \textbf{Video-level} & \multicolumn{3}{c}{\textbf{Phase-level}} \\
    \cmidrule(lr){3-3} \cmidrule(lr){4-4} \cmidrule(lr){5-7}
    & & \textbf{Accuracy $\uparrow$} & \textbf{Accuracy $\uparrow$} & \textbf{Precision $\uparrow$} & \textbf{Recall $\uparrow$} & \textbf{Jaccard $\uparrow$} \\
    \midrule
    \multirow{2}{*}{SelfSupSurg~\cite{selfsupsurg}} & \multirow{2}{*}{Cholec80~\cite{endonet}} &
    86.6 & 87.9 & 81.6 & 84.1 & 71.3 \\
&&(86.4, 86.8) & (87.7, 88.1) & (81.2, 82.0) & (83.7, 84.5) & (70.7, 71.9) \\
\midrule
    \multirow{2}{*}{EndoViT~\cite{endovit}} & \multirow{2}{*}{Endo700k~\cite{endovit}} & 
   87.1 & 88.4 & 81.9 & 84.3 & 71.7 \\
&&(86.9, 87.3) & (88.2, 88.6) & (81.5, 82.3) & (83.7, 84.9) & (71.1, 72.3) \\
\midrule
    \multirow{2}{*}{GSViT~\cite{GSViT}} &  \multirow{2}{*}{GenSurgery~\cite{GSViT}} & 
    69.2 & 72.2 & 66.5 & 72.7 & 51.9 \\
&&(69.0, 69.4) & (71.8, 72.6) & (65.9, 67.1) & (72.3, 73.1) & (51.5, 52.3) \\
\midrule
    \multirow{2}{*}{EndoSSL~\cite{endossl}} & \multirow{2}{*}{Lap-23M~\cite{endossl}} & 
   90.4 & 90.7 & 83.1 & 85.6 & 74.9 \\
&&(90.2, 90.6) & (90.5, 90.9) & (82.7, 83.5) & (85.2, 86.0) & (74.3, 75.5) \\
\midrule
     \multirow{2}{*}{SurgVISTA} & \multirow{2}{*}{SurgPub \& SurgWeb} & 
    \textbf{91.5} & \textbf{91.5} & \textbf{87.3} & \textbf{87.7} & \textbf{78.1} \\
    &&\textbf{(91.4, 91.7)} & \textbf{(91.4, 91.7)}& \textbf{(86.9, 87.7)} & \textbf{(87.4, 88.1)} & \textbf{(77.6, 78.6)} \\
    \midrule
    \textbf{P-value} & & $<$0.001 & $<$0.001 & $<$0.001 & $<$0.001 &$<$0.001 \\
    \bottomrule
	\end{tabular}}
    \label{lab:Cholec80_surgical}
\end{table*}
\begin{table*}[t]
    \centering
    \caption{Overall performance comparison between SurgVISTA and state-of-the-art surgical pre-training methods on the M2CAI16-Workflow dataset.}
   \resizebox{0.92\textwidth}{!}{
    \begin{tabular}	{ c c c c c c c c}
	\toprule
    \multirow{2}{*}{\textbf{Method}} & \multirow{2}{*}{\textbf{Pretraining Data}} & \textbf{Image-level} & \textbf{Video-level} & \multicolumn{3}{c}{\textbf{Phase-level}} \\
    \cmidrule(lr){3-3} \cmidrule(lr){4-4} \cmidrule(lr){5-7}
    & & \textbf{Accuracy $\uparrow$} & \textbf{Accuracy $\uparrow$} & \textbf{Precision $\uparrow$} & \textbf{Recall $\uparrow$} & \textbf{Jaccard $\uparrow$} \\
    \midrule
    \multirow{2}{*}{SelfSupSurg~\cite{selfsupsurg}} & \multirow{2}{*}{Cholec80~\cite{endonet}} &
84.2 & 85.2 & 80.9 & 82.6 & 68.7 \\
&&(83.8, 84.6) & (84.7, 85.6) & (79.9, 81.8) & (81.6, 83.7) & (67.7, 69.7) \\
\midrule
    \multirow{2}{*}{EndoViT~\cite{endovit}} & \multirow{2}{*}{Endo700k~\cite{endovit}} & 
   83.2 & 84.7 & 70.2 & 75.6 & 62.4 \\
&&(82.8, 83.6) & (84.3, 85.2) & (69.6, 70.8) & (75.1, 76.1) & (61.7, 63.1) \\
\midrule
    \multirow{2}{*}{GSViT~\cite{GSViT}} &  \multirow{2}{*}{GenSurgery~\cite{GSViT}} & 
    66.1 & 66.0 & 50.2 & 53.5 & 38.4 \\
&&(65.6, 66.7) & (65.4, 66.6) & (49.6, 50.9) & (52.9, 54.0) & (37.8, 39.0) \\
\midrule
    \multirow{2}{*}{EndoSSL~\cite{endossl}} & \multirow{2}{*}{Lap-23M~\cite{endossl}} & 
   86.1 & 87.2 & 72.2 & 77.1 & 65.7 \\
&&(85.7, 86.4) & (86.8, 87.6) & (71.6, 72.7) & (76.6, 77.6) & (65.1, 66.4) \\
\midrule
    \multirow{2}{*}{SurgVISTA} & \multirow{2}{*}{SurgPub \& SurgWeb} & 
    \textbf{87.2} & \textbf{88.9} & \textbf{83.5} & \textbf{86.2} & \textbf{73.2} \\
&&\textbf{(86.8, 87.6)} & \textbf{(88.5, 89.3)} & \textbf{(82.9, 84.0)} & \textbf{(85.6, 86.8)} & \textbf{(72.4, 74.0)} \\
\midrule
    \textbf{P-value} & & $<$0.001&$<$0.001&$<$0.001&$<$0.001&$<$0.001\\
    \bottomrule
	\end{tabular}}
    \label{lab:M2CAI16-Workflow_surgical}
\end{table*}
\begin{table*}[t]
    \centering
    \caption{Overall performance comparison between SurgVISTA and state-of-the-art surgical pre-training methods on the AutoLaparo dataset.}
    \resizebox{0.92\textwidth}{!}{
    \begin{tabular}	{ c c c c c c c c}
	\toprule
    \multirow{2}{*}{\textbf{Method}} & \multirow{2}{*}{\textbf{Pretraining Data}} & \textbf{Image-level} & \textbf{Video-level} & \multicolumn{3}{c}{\textbf{Phase-level}} \\
    \cmidrule(lr){3-3} \cmidrule(lr){4-4} \cmidrule(lr){5-7}
    & & \textbf{Accuracy $\uparrow$} & \textbf{Accuracy $\uparrow$} & \textbf{Precision $\uparrow$} & \textbf{Recall $\uparrow$} & \textbf{Jaccard $\uparrow$} \\
    \midrule
    \multirow{2}{*}{SelfSupSurg~\cite{selfsupsurg}} & \multirow{2}{*}{Cholec80~\cite{endonet}} &
77.4 & 78.2 & 57.3 & 61.4 & 50.4 \\
&&(76.9, 77.9) & (77.8, 78.7) & (57.2, 57.8) & (61.1, 61.7) & (50.0, 50.9) \\
\midrule
    \multirow{2}{*}{EndoViT~\cite{endovit}} & \multirow{2}{*}{Endo700k~\cite{endovit}} & 
   78.0 & 78.1 & 71.0 & 65.0 & 53.7 \\
&&(77.6, 78.5) & (77.7, 78.6) & (70.7, 71.4) & (64.4, 65.6) & (53.1, 54.4) \\
\midrule
    \multirow{2}{*}{GSViT~\cite{GSViT}} &  \multirow{2}{*}{GenSurgery~\cite{GSViT}} & 
    56.2 & 55.1 & 36.0 & 41.2 & 27.8 \\
&&(55.7, 56.8) & (54.6, 55.7) & (34.9, 37.1) & (39.5, 42.9) & (26.4, 29.2) \\
\midrule
    \multirow{2}{*}{EndoSSL~\cite{endossl}} & \multirow{2}{*}{Lap-23M~\cite{endossl}} & 
   85.3 & 85.6 & 75.2 & 70.5 & 61.7 \\
&&(84.9, 85.7) & (85.2, 86.0) & (74.9, 75.5) & (70.0, 71.0) & (61.1, 62.4) \\
\midrule
    \multirow{2}{*}{SurgVISTA} & \multirow{2}{*}{SurgPub \& SurgWeb} & 
    \textbf{87.5} & \textbf{87.8} & \textbf{86.6} & \textbf{75.3} & \textbf{66.8} \\
&&\textbf{(87.1, 87.8)} & \textbf{(87.4, 88.2)} & \textbf{(80.6, 92.7)} & \textbf{(74.4, 76.1)} & \textbf{(65.8, 67.7)} \\
\midrule
    \textbf{P-value} & & $<$0.001&$<$0.001&$<$0.001&$<$0.001&$<$0.001\\
    \bottomrule
	\end{tabular}}
    \label{lab:AutoLaparo_surgical}
\end{table*}
\begin{table*}[t]
    \centering
    \caption{Overall performance comparison between SurgVISTA and state-of-the-art surgical pre-training methods on the CATARACTS dataset.}
    \resizebox{0.92\textwidth}{!}{
    \begin{tabular}	{ c c c c c c c c}
	\toprule
    \multirow{2}{*}{\textbf{Method}} & \multirow{2}{*}{\textbf{Pretraining Data}} & \textbf{Image-level} & \textbf{Video-level} & \multicolumn{3}{c}{\textbf{Phase-level}} \\
    \cmidrule(lr){3-3} \cmidrule(lr){4-4} \cmidrule(lr){5-7}
    & & \textbf{Accuracy $\uparrow$} & \textbf{Accuracy $\uparrow$} & \textbf{Precision $\uparrow$} & \textbf{Recall $\uparrow$} & \textbf{Jaccard $\uparrow$} \\
    \midrule
    \multirow{2}{*}{SelfSupSurg~\cite{selfsupsurg}} & \multirow{2}{*}{Cholec80~\cite{endonet}} &
74.8 & 79.3 & 60.9 & 58.7 & 49.3 \\
&&(70.1, 75.5) & (78.6, 80.1) & (60.2, 61.6) & (57.8, 59.6) & (48.3, 50.3) \\
\midrule
    \multirow{2}{*}{EndoViT~\cite{endovit}} & \multirow{2}{*}{Endo700k~\cite{endovit}} & 
   76.8 & 80.9 & 58.1 & 52.9 & 46.1 \\
&&(76.1, 77.5) & (80.2, 81.6) & (57.4, 58.9) & (52.1, 53.7) & (45.2, 47.0) \\
\midrule
    \multirow{2}{*}{GSViT~\cite{GSViT}} &  \multirow{2}{*}{GenSurgery~\cite{GSViT}} & 
    53.0 & 57.0 & 24.9 & 26.8 & 19.2 \\
&&(52.1, 53.8) & (56.2, 57.9) & (24.5, 25.4) & (26.4, 27.3) & (18.7, 19.6) \\
\midrule
    \multirow{2}{*}{EndoSSL~\cite{endossl}} & \multirow{2}{*}{Lap-23M~\cite{endossl}} & 
   73.9 & 79.7 & 56.0 & 54.4 & 45.4 \\
&&(73.2, 74.6) & (79.0, 80.4) & (55.2, 56.8) & (53.7, 55.1) & (44.6, 46.2) \\
\midrule
    \multirow{2}{*}{SurgVISTA} & \multirow{2}{*}{SurgPub \& SurgWeb} & 
    \textbf{82.1} &\textbf{85.5} & \textbf{75.7} & \textbf{69.8} & \textbf{63.6} \\
&&\textbf{(81.5, 82.8)} & \textbf{(84.9, 86.2)} & \textbf{(75.0, 76.5)} & \textbf{(68.8, 70.9)} & \textbf{(62.4, 64.8)} \\
\midrule
    \textbf{P-value} & & $<$0.001&$<$0.001&$<$0.001&$<$0.001&$<$0.001\\
    \bottomrule
	\end{tabular}}
    \label{lab:CATARACTS_surgical}
\end{table*}

\begin{table*}[t]
    \centering
    \caption{Overall performance comparison between SurgVISTA and state-of-the-art surgical pre-training methods on the Cataract-101 dataset.}
    \resizebox{0.92\textwidth}{!}{
    \begin{tabular}	{ c c c c c c c c}
	\toprule
    \multirow{2}{*}{\textbf{Method}} & \multirow{2}{*}{\textbf{Pretraining Data}} & \textbf{Image-level} & \textbf{Video-level} & \multicolumn{3}{c}{\textbf{Phase-level}} \\
    \cmidrule(lr){3-3} \cmidrule(lr){4-4} \cmidrule(lr){5-7}
    & & \textbf{Accuracy $\uparrow$} & \textbf{Accuracy $\uparrow$} & \textbf{Precision $\uparrow$} & \textbf{Recall $\uparrow$} & \textbf{Jaccard $\uparrow$} \\
    \midrule
    \multirow{2}{*}{SelfSupSurg~\cite{selfsupsurg}} & \multirow{2}{*}{Cholec80~\cite{endonet}} &
86.1 & 87.5 & 85.7 & 85.1 & 74.5 \\
&&(85.6, 86.7) & (87.0, 88.0) & (85.0, 86.3) & (84.4, 85.8) & (73.6, 75.3) \\
\midrule
    \multirow{2}{*}{EndoViT~\cite{endovit}} & \multirow{2}{*}{Endo700k~\cite{endovit}} & 
   85.7 & 86.3 & 85.0 & 86.0 & 74.5 \\
&&(85.2, 86.2) & (85.8, 86.8) & (84.3, 85.6) & (85.3, 86.6) & (73.6, 75.4) \\
\midrule
    \multirow{2}{*}{GSViT~\cite{GSViT}} &  \multirow{2}{*}{GenSurgery~\cite{GSViT}} & 
    71.2 & 74.1 & 74.9 & 72.7 & 58.3 \\
&&(70.6, 71.8) & (73.5, 74.8) & (73.9, 75.9) & (71.8, 73.7) & (57.3, 59.4) \\
\midrule
    \multirow{2}{*}{EndoSSL~\cite{endossl}} & \multirow{2}{*}{Lap-23M~\cite{endossl}} & 
   86.3 & 88.0 & 86.4 & 87.4 & 76.5 \\
&&(85.9, 86.8) & (87.5, 88.5) & (85.8, 87.0) & (86.8, 88.0) & (75.6, 77.4) \\
\midrule
    \multirow{2}{*}{SurgVISTA} & \multirow{2}{*}{SurgPub \& SurgWeb} & 
    \textbf{94.2} & \textbf{94.8} & \textbf{92.9} & \textbf{93.8} & \textbf{87.6} \\
&&\textbf{(93.9, 94.6)} & \textbf{(94.4, 95.1)} & \textbf{(92.4, 93.4)} & \textbf{(93.4, 94.3)} & \textbf{(86.9, 88.4)} \\
\midrule
    \textbf{P-value} & & $<$0.001&$<$0.001&$<$0.001&$<$0.001&$<$0.001\\
    \bottomrule
	\end{tabular}}
    \label{lab:Cataract-101_surgical}
\end{table*}
\begin{table*}[t]
    \centering
    \caption{Overall performance comparison between SurgVISTA and state-of-the-art surgical pre-training methods on the Cataract-21 dataset.}
    \resizebox{0.92\textwidth}{!}{
    \begin{tabular}	{ c c c c c c c c}
	\toprule
    \multirow{2}{*}{\textbf{Method}} & \multirow{2}{*}{\textbf{Pretraining Data}} & \textbf{Image-level} & \textbf{Video-level} & \multicolumn{3}{c}{\textbf{Phase-level}} \\
    \cmidrule(lr){3-3} \cmidrule(lr){4-4} \cmidrule(lr){5-7}
    & & \textbf{Accuracy $\uparrow$} & \textbf{Accuracy $\uparrow$} & \textbf{Precision $\uparrow$} & \textbf{Recall $\uparrow$} & \textbf{Jaccard $\uparrow$} \\
    \midrule
    \multirow{2}{*}{SelfSupSurg~\cite{selfsupsurg}} & \multirow{2}{*}{Cholec80~\cite{endonet}} &
88.3 & 88.8 & 85.1 & 84.3 & 73.4 \\
&&(86.3, 90.3) & (86.8, 90.7) & (82.0, 88.2) & (81.5, 87.1) & (69.5, 77.3) \\
\midrule
    \multirow{2}{*}{EndoViT~\cite{endovit}} & \multirow{2}{*}{Endo700k~\cite{endovit}} & 
   87.0 & 86.6 & 78.0 & 75.1 & 65.9 \\
&&(84.8, 89.2) & (84.3, 88.9) & (75.8, 80.2) & (72.5, 77.6) & (62.6, 69.2) \\
\midrule
    \multirow{2}{*}{GSViT~\cite{GSViT}} &  \multirow{2}{*}{GenSurgery~\cite{GSViT}} & 
    72.3 & 73.2 & 60.5 & 71.3 & 52.1 \\
&&(69.5, 75.1) & (70.4, 76.1) & (57.1, 63.9) & (67.7, 74.9) & (48.1, 56.1) \\
\midrule
    \multirow{2}{*}{EndoSSL~\cite{endossl}} & \multirow{2}{*}{Lap-23M~\cite{endossl}} & 
   84.9 & 83.7 & 85.8 & 78.9 & 69.0 \\
&&(82.6, 87.1) & (81.2, 86.2) & (81.3, 90.3) & (73.6, 84.2) & (63.2, 74.9) \\
\midrule
    \multirow{2}{*}{SurgVISTA} & \multirow{2}{*}{SurgPub \& SurgWeb} & 
    \textbf{92.1} & \textbf{92.0} & \textbf{89.8} & \textbf{88.7} & \textbf{80.0} \\
&&\textbf{(90.4, 93.8)} & \textbf{(90.2, 93.8)} & \textbf{(87.2, 92.5)} & \textbf{(85.9, 91.4)} & \textbf{(75.9, 84.2)} \\
\midrule
    \textbf{P-value} & &$<$0.001&$<$0.001&$<$0.001&$<$0.001&$<$0.001\\
    \bottomrule
	\end{tabular}}
    \label{lab:Cataract-21_surgical}
\end{table*}
\begin{table*}[t]
    \centering
    \caption{Overall comparison under different data configurations on the Cholec80 dataset. \textcolor{red}{Red} denotes the volume of data corresponding to the target surgical procedure, while \textcolor{blue}{Blue} indicates the total data volume utilized in each experimental setting.}
    \resizebox{0.92\textwidth}{!}{
    \begin{tabular}	{ c c c c c c c c}
	\toprule
    \multirow{2}{*}{\textbf{Setting}} & \multirow{2}{*}{\textbf{Pretraining Data}} & \textbf{Image-level} & \textbf{Video-level} & \multicolumn{3}{c}{\textbf{Phase-level}} \\
    \cmidrule(lr){3-3} \cmidrule(lr){4-4} \cmidrule(lr){5-7}
    & & \textbf{Accuracy $\uparrow$} & \textbf{Accuracy $\uparrow$} & \textbf{Precision $\uparrow$} & \textbf{Recall $\uparrow$} & \textbf{Jaccard $\uparrow$} \\
    \midrule
    \multirow{2}{*}{Setting A} & \multirow{2}{*}{\textcolor{red}{86344} / \textcolor{blue}{86344}} &
   89.4 & 89.5 & 85.4 & 83.7 & 72.8 \\
&&(89.2, 89.6) & (89.3, 89.7) & (85.0, 85.8) & (83.4, 84.1) & (72.3, 73.3) \\
\midrule
    \multirow{2}{*}{Setting B} & \multirow{2}{*}{\textcolor{red}{153922} / \textcolor{blue}{153922}} & 
   90.7 & 90.9 & 86.8 & 86.8 & 76.6 \\
&&(90.5, 90.9) & (90.7, 91.1) & (86.5, 87.2) & (86.5, 87.2) & (76.1, 77.1) \\
\midrule
    \multirow{2}{*}{Setting C} &  \multirow{2}{*}{\textcolor{red}{209061} / \textcolor{blue}{441608}} & 
    91.1 & 91.3 & 87.1 & 87.5 & 77.7 \\
&&(90.9, 91.3) & (91.1, 91.5) & (86.6, 87.6) & (87.0, 88.0) & (77.1, 78.3) \\
\midrule
    \multirow{2}{*}{Setting D} & \multirow{2}{*}{\textcolor{red}{209061} / \textcolor{blue}{1203159}} & 
   91.2 & 91.4 & \textbf{87.4} & \textbf{88.0} & 77.7 \\
&&(91.0, 91.3) & (91.2, 91.6) & \textbf{(87.0, 87.7)} & \textbf{(87.6, 88.3)} & (77.2, 78.2) \\
\midrule
    \multirow{2}{*}{Setting E} & \multirow{2}{*}{\textcolor{red}{535295} / \textcolor{blue}{3552777}} & 
    \textbf{91.5} & \textbf{91.5} & 87.3 & 87.7 & \textbf{78.1} \\
    &&\textbf{(91.4, 91.7)} & \textbf{(91.4, 91.7)}& (86.9, 87.7) & (87.4, 88.1) & \textbf{(77.6, 78.6)} \\
    \bottomrule
	\end{tabular}}
    \label{lab:Cholec80_scalinglaw}
\end{table*}
\begin{table*}[t]
    \centering
    \caption{Overall comparison under different data configurations on the M2CAI16-Workflow dataset. \textcolor{red}{Red} denotes the volume of data corresponding to the target surgical procedure, while \textcolor{blue}{Blue} indicates the total data volume utilized in each experimental setting.}
    \resizebox{0.92\textwidth}{!}{
    \begin{tabular}	{ c c c c c c c c}
	\toprule
    \multirow{2}{*}{\textbf{Setting}} & \multirow{2}{*}{\textbf{Pretraining Data}} & \textbf{Image-level} & \textbf{Video-level} & \multicolumn{3}{c}{\textbf{Phase-level}} \\
    \cmidrule(lr){3-3} \cmidrule(lr){4-4} \cmidrule(lr){5-7}
    & & \textbf{Accuracy $\uparrow$} & \textbf{Accuracy $\uparrow$} & \textbf{Precision $\uparrow$} & \textbf{Recall $\uparrow$} & \textbf{Jaccard $\uparrow$} \\
    \midrule
    \multirow{2}{*}{Setting A} & \multirow{2}{*}{\textcolor{red}{86344} / \textcolor{blue}{86344}} &
   83.9 & 85.7 & 82.1 & 82.6 & 69.0 \\
&&(83.6, 84.5) & (85.3, 86.1) & (81.5, 82.7) & (82.0, 83.2) & (68.2, 69.7) \\
\midrule
    \multirow{2}{*}{Setting B} & \multirow{2}{*}{\textcolor{red}{153922} / \textcolor{blue}{153922}} & 
   84.9 & 86.9 & 83.3 & 82.6 & 70.2 \\
&&(84.7, 85.5) & (86.5, 87.3) & (82.7, 83.9) & (81.9, 83.3) & (69.4, 71.0) \\
\midrule
    \multirow{2}{*}{Setting C} &  \multirow{2}{*}{\textcolor{red}{209061} / \textcolor{blue}{441608}} & 
    85.0 & 86.9 & 83.1 & 84.0 & 71.4 \\
&&(84.6, 85.4) & (86.5, 87.3) & (82.6, 83.7) & (83.4, 84.7) & (70.7, 72.2) \\
\midrule
    \multirow{2}{*}{Setting D} & \multirow{2}{*}{\textcolor{red}{209061} / \textcolor{blue}{1203159}} & 
   86.6 & 88.3 & \textbf{84.6} & 84.2 & 73.5 \\
&&(86.3, 87.0) & (87.9, 88.7) & \textbf{(83.9, 85.4)} & (83.4, 85.0) & (72.6, 74.3) \\
\midrule
    \multirow{2}{*}{Setting E} & \multirow{2}{*}{\textcolor{red}{535295} / \textcolor{blue}{3552777}} & 
    \textbf{87.2} & \textbf{88.9} & 83.5 & \textbf{86.2} & \textbf{73.2} \\
&&\textbf{(86.8, 87.6)} & \textbf{(88.5, 89.3)} & (82.9, 84.0) & \textbf{(85.6, 86.8)} & \textbf{(72.4, 74.0)} \\
    \bottomrule
	\end{tabular}}
    \label{lab:M2CAI16_scalinglaw}
\end{table*}
\begin{table*}[t]
    \centering
    \caption{Overall comparison under different data configurations on the CholecT50 dataset. \textcolor{red}{Red} denotes the volume of data corresponding to the target surgical procedure, while \textcolor{blue}{Blue} indicates the total data volume utilized in each experimental setting.}
    \resizebox{0.92\textwidth}{!}{
    \begin{tabular}	{ c c c c c c c c c}
	\toprule
    \multirow{2}{*}{\textbf{Setting}} & \multirow{2}{*}{\textbf{Pretraining Data}} & \multicolumn{3}{c}{\textbf{Component detection}} & \multicolumn{3}{c}{\textbf{Triplet association}} \\
    \cmidrule(lr){3-5} \cmidrule(lr){6-8}
    & & \bm{$AP_I$} & \bm{$AP_v$} & \bm{$AP_t$} & \bm{$AP_{iv}$} & \bm{$AP_{it}$} & \bm{$AP_{ivt}$} \\
    \midrule
    \multirow{2}{*}{Setting A} & \multirow{2}{*}{\textcolor{red}{86344} / \textcolor{blue}{86344}} &
  81.4 & 51.5 & 44.9 & 35.0 & 40.2 & 35.5 \\
&&(79.7, 83.0) & (49.4, 53.6) & (41.2, 48.5) & (32.2, 37.7) & (37.7, 42.7) & (33.2, 37.8) \\
\midrule
    \multirow{2}{*}{Setting B} & \multirow{2}{*}{\textcolor{red}{153922} / \textcolor{blue}{153922}} & 
   83.0 & 54.0 & 48.2 & 37.9 & 44.2 & 39.5 \\
&&(82.0, 84.1) & (51.8, 56.2) & (44.3, 52.1) & (34.6, 41.2) & (41.4, 47.1) & (36.8, 42.3) \\
\midrule
    \multirow{2}{*}{Setting C} &  \multirow{2}{*}{\textcolor{red}{209061} / \textcolor{blue}{441608}} & 
    83.0 & 54.0 & 48.2 & 37.9 & 44.2 & 39.5 \\
&&(82.0, 84.1) & (51.8, 56.2) & (44.3, 52.1) & (34.6, 41.2) & (41.4, 47.1) & (36.8, 42.3) \\
\midrule
    \multirow{2}{*}{Setting D} & \multirow{2}{*}{\textcolor{red}{209061} / \textcolor{blue}{1203159}} & 
   \textbf{88.01} & \textbf{57.89} & 50.10 & 41.44 & 42.98 & 39.47 \\
&&\textbf{(86.8, 89.2)} & \textbf{(55.2, 60.6)} & (45.9, 54.3) & (38.1, 44.8) & (40.4, 45.6) & (36.6, 42.3) \\
\midrule
    \multirow{2}{*}{Setting E} & \multirow{2}{*}{\textcolor{red}{535295} / \textcolor{blue}{3552777}} & 
    84.3 & 53.7 & \textbf{51.6} & \textbf{41.6} & \textbf{44.3} & \textbf{41.1} \\
&&(83.1, 85.6) & (50.6, 56.9) & \textbf{(47.3, 55.9)} & \textbf{(37.2, 46.0)} & \textbf{(40.9, 47.5)} & \textbf{(38.0, 44.1)} \\
    \bottomrule
\end{tabular}}
    \label{lab:CholecT50_scalinglaw}
\end{table*}
\begin{table*}[t]
    \centering
    \caption{Overall comparison under different data configurations on the PmLR50 dataset. \textcolor{red}{Red} denotes the volume of data corresponding to the target surgical procedure, while \textcolor{blue}{Blue} indicates the total data volume utilized in each experimental setting.}
    \resizebox{0.92\textwidth}{!}{
    \begin{tabular}	{ c c c c c c c c}
	\toprule
    \multirow{2}{*}{\textbf{Setting}} & \multirow{2}{*}{\textbf{Pretraining Data}} & \textbf{Image-level} & \textbf{Video-level} & \multicolumn{3}{c}{\textbf{Phase-level}} \\
    \cmidrule(lr){3-3} \cmidrule(lr){4-4} \cmidrule(lr){5-7}
    & & \textbf{Accuracy $\uparrow$} & \textbf{Accuracy $\uparrow$} & \textbf{Precision $\uparrow$} & \textbf{Recall $\uparrow$} & \textbf{Jaccard $\uparrow$} \\
    \midrule
    \multirow{2}{*}{Setting A} & \multirow{2}{*}{\textcolor{red}{0} / \textcolor{blue}{86344}} &
   88.5 & 87.9 & 85.1 & 75.3 & 66.3 \\
&&(87.7, 89.3) & (87.1, 88.8) & (83.3, 86.8) & (73.6, 77.1) & (64.3, 68.4) \\
\midrule
    \multirow{2}{*}{Setting B} & \multirow{2}{*}{\textcolor{red}{0} / \textcolor{blue}{153922}} & 
   89.9 & 89.4 & 87.4 & 79.8 & 71.5 \\
&&(89.2, 90.7) & (88.6, 90.2) & (85.6, 89.2) & (78.2, 81.5) & (69.4, 73.5) \\
\midrule
    \multirow{2}{*}{Setting C} &  \multirow{2}{*}{\textcolor{red}{0} / \textcolor{blue}{441608}} & 
    90.3 & 89.6 & 88.2 & 82.6 & 74.3 \\
&&(89.5, 91.0) & (88.8, 90.4) & (86.5, 89.9) & (80.7, 84.4) & (72.2, 76.5) \\
\midrule
    \multirow{2}{*}{Setting D} & \multirow{2}{*}{\textcolor{red}{0} / \textcolor{blue}{1203159}} & 
   91.9 & 91.3 & 87.8 & 84.9 & 76.4 \\
&&(91.2, 92.6) & (90.6, 92.1) & (86.2, 89.3) & (83.6, 86.2) & (74.7, 78.1) \\
\midrule
    \multirow{2}{*}{Setting E} & \multirow{2}{*}{\textcolor{red}{67779} / \textcolor{blue}{3552777}} & 
    \textbf{93.2} & \textbf{92.6} & \textbf{92.8} & \textbf{90.0} & \textbf{83.5} \\
&&\textbf{(92.6, 93.8)} & \textbf{(92.0, 93.3)} & \textbf{(91.8, 93.9)} & \textbf{(88.4, 91.7)} & \textbf{(81.6, 85.5)} \\
    \bottomrule
	\end{tabular}}
    \label{lab:PmLR50_scalinglaw}
\end{table*}
\begin{table*}[t]
    \centering
    \caption{Overall comparison under different data configurations on the AutoLaparo dataset. \textcolor{red}{Red} denotes the volume of data corresponding to the target surgical procedure, while \textcolor{blue}{Blue} indicates the total data volume utilized in each experimental setting.}
    \resizebox{0.92\textwidth}{!}{
    \begin{tabular}	{ c c c c c c c c}
	\toprule
    \multirow{2}{*}{\textbf{Setting}} & \multirow{2}{*}{\textbf{Pretraining Data}} & \textbf{Image-level} & \textbf{Video-level} & \multicolumn{3}{c}{\textbf{Phase-level}} \\
    \cmidrule(lr){3-3} \cmidrule(lr){4-4} \cmidrule(lr){5-7}
    & & \textbf{Accuracy $\uparrow$} & \textbf{Accuracy $\uparrow$} & \textbf{Precision $\uparrow$} & \textbf{Recall $\uparrow$} & \textbf{Jaccard $\uparrow$} \\
    \midrule
    \multirow{2}{*}{Setting A} & \multirow{2}{*}{\textcolor{red}{0} / \textcolor{blue}{86344}} &
   81.7 & 82.2 & 73.7 & 69.8 & 59.7 \\
&&(81.3, 82.1) & (81.8, 82.7) & (72.4, 75.0) & (69.2, 70.5) & (59.0, 60.4) \\
\midrule
    \multirow{2}{*}{Setting B} & \multirow{2}{*}{\textcolor{red}{0} / \textcolor{blue}{153922}} & 
81.3 & 81.7 & 75.5 & 69.6 & 59.1 \\
&&(80.9, 81.7) & (81.3, 82.1) & (71.5, 79.5) & (69.0, 70.2) & (58.4, 59.8) \\
\midrule
    \multirow{2}{*}{Setting C} &  \multirow{2}{*}{\textcolor{red}{40211} / \textcolor{blue}{441608}} & 
    83.0 & 83.5 & 80.7 & 73.9 & 63.6 \\
&&(82.6, 83.4) & (83.1, 83.9) & (78.0, 83.4) & (72.6, 75.2) & (62.4, 64.8) \\
\midrule
    \multirow{2}{*}{Setting D} & \multirow{2}{*}{\textcolor{red}{40211} / \textcolor{blue}{1203159}} & 
   85.4 & 85.7 & 81.4 & \textbf{75.5} & 65.8 \\
&&(84.9, 85.8) & (85.3, 86.1) & (79.9, 83.0) & \textbf{(74.4, 76.6)} & (64.8, 66.9) \\
\midrule
    \multirow{2}{*}{Setting E} & \multirow{2}{*}{\textcolor{red}{40211} / \textcolor{blue}{3552777}} & 
    \textbf{87.5} & \textbf{87.8} & \textbf{86.6} & 75.3 & \textbf{66.8} \\
&&\textbf{(87.1, 87.8)} & \textbf{(87.4, 88.2)} & \textbf{(80.6, 92.7)} & (74.4, 76.1) & \textbf{(65.8, 67.7)} \\
    \bottomrule
	\end{tabular}}
    \label{lab:AutoLaparo_scalinglaw}
\end{table*}
\begin{table*}[t]
    \centering
    \caption{Overall comparison under different data configurations on the SurgicalActions160 dataset. \textcolor{red}{Red} denotes the volume of data corresponding to the target surgical procedure, while \textcolor{blue}{Blue} indicates the total data volume utilized in each experimental setting.}
    \resizebox{0.92\textwidth}{!}{
    \begin{tabular}	{ c c c c c c}
	\toprule
    \textbf{Setting} & \textbf{Setting A} & \textbf{Setting B} & \textbf{Setting C} & \textbf{Setting D} & \textbf{Setting E} \\
    \midrule
    \textbf{Pre-training} & \multirow{2}{*}{\textcolor{red}{0} / \textcolor{blue}{86344}} & \multirow{2}{*}{\textcolor{red}{0} / \textcolor{blue}{153922}} & \multirow{2}{*}{\textcolor{red}{40211} / \textcolor{blue}{441608}} & \multirow{2}{*}{\textcolor{red}{40211} / \textcolor{blue}{1203159}} & \multirow{2}{*}{\textcolor{red}{40211} / \textcolor{blue}{3552777}} \\
    \textbf{Data} &&&&& \\ 
    \midrule
    \multirow{2}{*}{\textbf{Acc} $\uparrow$} &
  39.3 & 41.7 & 48.0 & 52.1 & \textbf{66.7} \\
&(25.4, 53.1) & (27.6, 55.9) & (34.1, 61.9) & (37.5, 66.6) & \textbf{(53.2, 80.2)} \\
\midrule
    \multirow{2}{*}{\textbf{mAP} $\uparrow$} & 
   50.0 & 48.3 & 58.2 & 62.3 & \textbf{67.6} \\
&(36.5, 63.6) & (35.5, 61.1) & (44.9, 71.6) & (49.6, 75.1) & \textbf{(55.2, 80.0)} \\
    \bottomrule
	\end{tabular}}
    \label{lab:SurgicalActions160_scalinglaw}
\end{table*}
\begin{table*}[t]
    \centering
    \caption{Overall comparison under different data configurations on the CATARACTS dataset. \textcolor{red}{Red} denotes the volume of data corresponding to the target surgical procedure, while \textcolor{blue}{Blue} indicates the total data volume utilized in each experimental setting.}

    \resizebox{0.92\textwidth}{!}{
    \begin{tabular}	{ c c c c c c c c}
	\toprule
    \multirow{2}{*}{\textbf{Setting}} & \multirow{2}{*}{\textbf{Pretraining Data}} & \textbf{Image-level} & \textbf{Video-level} & \multicolumn{3}{c}{\textbf{Phase-level}} \\
    \cmidrule(lr){3-3} \cmidrule(lr){4-4} \cmidrule(lr){5-7}
    & & \textbf{Accuracy $\uparrow$} & \textbf{Accuracy $\uparrow$} & \textbf{Precision $\uparrow$} & \textbf{Recall $\uparrow$} & \textbf{Jaccard $\uparrow$} \\
    \midrule
    \multirow{2}{*}{Setting A} & \multirow{2}{*}{\textcolor{red}{0} / \textcolor{blue}{86344}} &
   79.4 & 83.5 & 73.1 & 63.5 & 55.9 \\
&&(78.7, 80.1) & (82.9, 84.2) & (72.1, 74.2) & (62.1, 64.9) & (54.6, 57.3) \\
\midrule
    \multirow{2}{*}{Setting B} & \multirow{2}{*}{\textcolor{red}{0} / \textcolor{blue}{153922}} & 
   80.0 & 84.1 & 72.4 & 66.7 & 58.9 \\
&&(79.3, 80.7) & (83.4, 84.7) & (71.4, 73.4) & (65.4, 67.9) & (57.6, 60.3) \\
\midrule
    \multirow{2}{*}{Setting C} &  \multirow{2}{*}{\textcolor{red}{0} / \textcolor{blue}{441608}} & 
    80.6 & 84.0 & 73.1 & 67.8 & 60.5 \\
&&(79.9, 81.2) & (83.4, 84.6) & (72.2, 74.0) & (66.5, 69.1) & (59.1, 61.8) \\
\midrule
    \multirow{2}{*}{Setting D} & \multirow{2}{*}{\textcolor{red}{0} / \textcolor{blue}{1203159}} & 
   80.9 & 84.4 & 71.8 & 69.8 & 61.1 \\
&&(80.2, 81.5) & (83.7, 85.0) & (70.9, 72.7) & (68.7, 70.9) & (59.87, 62.28) \\
\midrule
    \multirow{2}{*}{Setting E} & \multirow{2}{*}{\textcolor{red}{0} / \textcolor{blue}{3552777}} & 
   \textbf{82.1} & \textbf{85.5} & \textbf{75.7} & \textbf{69.8} & \textbf{63.6} \\
&&\textbf{(81.5, 82.8)} & \textbf{(84.9, 86.2)} & \textbf{(75.0, 76.5)} & \textbf{(68.8, 70.9)} & \textbf{(62.4, 64.8)} \\
    \bottomrule
	\end{tabular}}
    \label{lab:CATARACTS_scalinglaw}
\end{table*}
\begin{table*}[t]
    \centering
    \caption{Overall comparison under different data configurations on the Cataract-101 dataset. \textcolor{red}{Red} denotes the volume of data corresponding to the target surgical procedure, while \textcolor{blue}{Blue} indicates the total data volume utilized in each experimental setting.}
    \resizebox{0.92\textwidth}{!}{
    \begin{tabular}	{ c c c c c c c c}
	\toprule
    \multirow{2}{*}{\textbf{Setting}} & \multirow{2}{*}{\textbf{Pretraining Data}} & \textbf{Image-level} & \textbf{Video-level} & \multicolumn{3}{c}{\textbf{Phase-level}} \\
    \cmidrule(lr){3-3} \cmidrule(lr){4-4} \cmidrule(lr){5-7}
    & & \textbf{Accuracy $\uparrow$} & \textbf{Accuracy $\uparrow$} & \textbf{Precision $\uparrow$} & \textbf{Recall $\uparrow$} & \textbf{Jaccard $\uparrow$} \\
    \midrule
    \multirow{2}{*}{Setting A} & \multirow{2}{*}{\textcolor{red}{0} / \textcolor{blue}{86344}} &
   91.1 & 92.0 & 89.9 & 90.6 & 82.4 \\
&&(90.6, 91.5) & (91.6, 92.4) & (89.3, 90.5) & (90.1, 91.2) & (81.6, 83.3) \\
\midrule
    \multirow{2}{*}{Setting B} & \multirow{2}{*}{\textcolor{red}{0} / \textcolor{blue}{153922}} & 
   91.2 & 92.2 & 90.0 & 91.3 & 83.0 \\
&&(90.8, 91.6) & (91.8, 92.6) & (89.4, 90.6) & (90.8, 91.8) & (82.2, 83.8) \\
\midrule
    \multirow{2}{*}{Setting C} &  \multirow{2}{*}{\textcolor{red}{0} / \textcolor{blue}{441608}} & 
    92.3 & 93.1 & 90.6 & 91.7 & 84.1 \\
&&(91.9, 92.7) & (92.7, 93.4) & (90.1, 91.2) & (91.2, 92.2) & (83.3, 84.9) \\
\midrule
    \multirow{2}{*}{Setting D} & \multirow{2}{*}{\textcolor{red}{0} / \textcolor{blue}{1203159}} & 
   93.3 & 93.9 & 91.8 & 92.6 & 85.8 \\
&&(93.0, 93.7) & (93.5, 94.3) & (91.3, 92.4) & (92.1, 93.1) & (85.0, 86.6) \\
\midrule
    \multirow{2}{*}{Setting E} & \multirow{2}{*}{\textcolor{red}{0} / \textcolor{blue}{3552777}} & 
   \textbf{94.2} & \textbf{94.8} & \textbf{92.9} & \textbf{93.8} & \textbf{87.6} \\
&&\textbf{(93.9, 94.6)} & \textbf{(94.4, 95.1)} & \textbf{(92.4, 93.4)} & \textbf{(93.4, 94.3)} & \textbf{(86.9, 88.4)} \\
    \bottomrule
	\end{tabular}}
    \label{lab:Cataract-101_scalinglaw}
\end{table*}
\begin{table*}[t]
    \centering
    \caption{Overall comparison under different data configurations on the Cataract-21 dataset. \textcolor{red}{Red} denotes the volume of data corresponding to the target surgical procedure, while \textcolor{blue}{Blue} indicates the total data volume utilized in each experimental setting.}
    
    \resizebox{0.92\textwidth}{!}{
    \begin{tabular}	{ c c c c c c c c}
	\toprule
    \multirow{2}{*}{\textbf{Setting}} & \multirow{2}{*}{\textbf{Pretraining Data}} & \textbf{Image-level} & \textbf{Video-level} & \multicolumn{3}{c}{\textbf{Phase-level}} \\
    \cmidrule(lr){3-3} \cmidrule(lr){4-4} \cmidrule(lr){5-7}
    & & \textbf{Accuracy $\uparrow$} & \textbf{Accuracy $\uparrow$} & \textbf{Precision $\uparrow$} & \textbf{Recall $\uparrow$} & \textbf{Jaccard $\uparrow$} \\
    \midrule
    \multirow{2}{*}{Setting A} & \multirow{2}{*}{\textcolor{red}{0} / \textcolor{blue}{86344}} &
   89.4 & 89.0 & 84.7 & 85.3 & 74.1 \\
&&(87.4, 91.4) & (86.9, 91.1) & (81.6, 87.8) & (82.2, 88.4) & (69.7, 78.4) \\
\midrule
    \multirow{2}{*}{Setting B} & \multirow{2}{*}{\textcolor{red}{0} / \textcolor{blue}{153922}} & 
   90.1 & 89.8 & 87.0 & 87.4 & 77.8 \\
&&(88.2, 91.9) & (87.9, 91.7) & (84.4, 89.7) & (85.0, 89.8) & (74.1, 81.5) \\
\midrule
    \multirow{2}{*}{Setting C} &  \multirow{2}{*}{\textcolor{red}{0} / \textcolor{blue}{441608}} & 
    90.0 & 89.7 & 85.6 & 86.4 & 75.8 \\
&&(88.1, 91.8) & (87.7, 91.6) & (82.7, 88.6) & (83.9, 88.8) & (71.9, 79.6) \\
\midrule
    \multirow{2}{*}{Setting D} & \multirow{2}{*}{\textcolor{red}{0} / \textcolor{blue}{1203159}} & 
   90.6 & 90.4 & 84.9 & 86.8 & 75.7 \\
&&(88.7, 92.4) & (88.5, 92.3) & (81.8, 87.9) & (84.2, 89.4) & (71.9, 79.6) \\
\midrule
    \multirow{2}{*}{Setting E} & \multirow{2}{*}{\textcolor{red}{0} / \textcolor{blue}{3552777}} & 
   \textbf{92.1} & \textbf{92.0} & \textbf{89.8} & \textbf{88.7} & \textbf{80.0} \\
&&\textbf{(90.4, 93.8)} & \textbf{(90.2, 93.8)} & \textbf{(87.2, 92.5)} & \textbf{(85.9, 91.4)} & \textbf{(75.9, 84.2)} \\
    \bottomrule
	\end{tabular}}
    \label{lab:Cataract-21_scalinglaw}
\end{table*}
\begin{table}
\centering
\caption{Detailed configurations of the natural-domain pre-trained methods employed in this study.}
\begin{tabular}{l|l|l|p{9cm}} 
\toprule
\textbf{Dataset} & \textbf{Method} & \textbf{Backbone} & \textbf{URL} \\
\midrule
ImageNet-1K~\cite{imagenet} & Supervised~\cite{ViT} & ViT-B & \url{https://huggingface.co/timm/vit_base_patch16_224.augreg_in1k} \\
\midrule
ImageNet-1K~\cite{imagenet} & DINO~\cite{DINO} & ViT-B & \url{https://huggingface.co/timm/vit_base_patch16_224.dino} \\
\midrule
ImageNet-1K~\cite{imagenet,MAE} & MAE~\cite{MAE} & ViT-B & \url{https://huggingface.co/timm/vit_base_patch16_224.mae} \\
\midrule
Kinetics-400~\cite{k400} & VideoMAE~\cite{videomae} & ViT-B & \url{https://github.com/MCG-NJU/VideoMAE} \\
\midrule
UnlabeledHybrid~\cite{internvideo} & VideoMAE~\cite{videomae} & ViT-B & \url{https://github.com/OpenGVLab/InternVideo} \\
\midrule
SSV2~\cite{ssv2} & VideoMAE~\cite{videomae} & ViT-B & \url{https://github.com/MCG-NJU/VideoMAE} \\
\midrule
Kinetics-710~\cite{k710} & UMT~\cite{UMT} & ViT-B & \url{https://huggingface.co/OpenGVLab/UMT} \\
\midrule
Kinetics-400~\cite{k400} & MVD~\cite{MVD} & ViT-B & \url{https://github.com/ruiwang2021/mvd} \\
\bottomrule
\end{tabular}
\label{tab:natural_params}
\end{table}
\begin{table*}
\centering
\caption{Detailed descriptions of the surgical-domain pre-trained methods employed in this study.}
\begin{tabular}{l|l|l|l|p{7.2cm}} 
\toprule
\textbf{Method} & \textbf{Dataset} & \textbf{Method} & \textbf{Backbone} & \textbf{URL} \\
\midrule
SelfSupSurg~\cite{selfsupsurg} & Cholec80~\cite{endonet} & DINO~\cite{DINO} & ResNet-50~\cite{resnet} & \url{https://github.com/CAMMA-public/SelfSupSurg} \\
\midrule
EndoSSL~\cite{endossl}  & Private & MSN~\cite{MSN} & ViT-L~\cite{ViT} & \url{https://github.com/RoyHirsch/endossl} \\
\midrule
EndoViT~\cite{endovit} &  Endo700k~\cite{endovit}  & MAE~\cite{MAE}  & ViT-B~\cite{ViT} & \url{https://github.com/DominikBatic/EndoViT} \\
\midrule
GSViT~\cite{GSViT}  &  GenSurgery~\cite{GSViT} & GSViT & EfficientNet~\cite{efficientvit} & \url{https://github.com/SamuelSchmidgall/GSViT} \\
\bottomrule
\end{tabular}
\label{tab:surgical_params}
\end{table*}
\begin{table*}[ht]
\centering
\caption{The public codes used in this study.}
\begin{tabular}{l|p{7.5cm}}
\toprule
\textbf{Name} & \textbf{URL} \\
\midrule
VideoMAE~\cite{videomae} & \href{https://github.com/MCG-NJU/VideoMAE}{https://github.com/MCG-NJU/VideoMAE} \\
\midrule
MVD~\cite{MVD} & \href{https://github.com/ruiwang2021/mvd}{https://github.com/ruiwang2021/mvd} \\
\midrule
TeCNO~\cite{TeCNO} & \href{https://github.com/tobiascz/TeCNO}{https://github.com/tobiascz/TeCNO} \\
\midrule
ivtmetrics~\cite{CholecT50_split} & \href{https://github.com/CAMMA-public/ivtmetrics}{https://github.com/CAMMA-public/ivtmetrics} \\
\midrule
Surgformer~\cite{yang2024surgformer} & \href{https://github.com/isyangshu/Surgformer}{https://github.com/isyangshu/Surgformer} \\
\midrule
TMRNet~\cite{TMRNet} & \href{https://github.com/YuemingJin/TMRNet}{https://github.com/YuemingJin/TMRNet} \\

\bottomrule
\end{tabular}
\label{tab:public_code} 
\end{table*}
\begin{table*}[!ht]
\centering
\caption{Data availability of public datasets used in the pre-training and fine-tuning phases. ``Open Access'' datasets are publicly available without restrictions. ``Credentialed Access'' datasets  require user credentials, such as account registration, email verification, identity confirmation, or participation in an affiliated competition. “License-Based Access” datasets require users to agree to specific licensing terms prior to use.}
\adjustbox{width=1\linewidth}{
\begin{tabular}{c|>{\centering\arraybackslash}p{11cm}|c}
\toprule
 \textbf{Dataset Name} &  \textbf{Link} &  \textbf{Access} \\
\midrule
\rowcolor{dino}
\multicolumn{3}{c}{\textbf{Laparoscopic Cholecystectomy} } \\
\midrule
Cholec80~\cite{endonet}&\url{https://camma.unistra.fr/datasets/} &  Credentialed Access \\
\midrule
Cholec80-CVS~\cite{Cholec80-CVS}&\url{https://github.com/ManuelRios18/CHOLEC80-CVS-PUBLIC} &  Open Access \\
\midrule
CholecT50~\cite{CholecT50}&\url{https://github.com/CAMMA-public/cholect50} &  Credentialed Access \\
\midrule
M2CAI16-Workflow~\cite{endonet}&\url{https://camma.unistra.fr/m2cai2016/index.php/program-challenge/} &  Credentialed Access \\
\midrule
HeiChole~\cite{HeiChole}&\url{https://www.synapse.org/Synapse:syn18824884/wiki/592580} &  Credentialed Access \\
\midrule
Endoscapes-CVS~\cite{Endoscapes-CVS201} &\url{https://github.com/CAMMA-public/Endoscapes} &  Open Access \\
\hline
\end{tabular}
}
\label{data_available_training}
\end{table*}
\end{appendices}

\end{document}